\documentclass[letterpaper,journal]{IEEEtran}
\usepackage{amsmath,amsfonts}
\usepackage{algorithmic}
\usepackage{algorithm}
\usepackage{array}
\usepackage[caption=false,font=normalsize,labelfont=sf,textfont=sf]{subfig}
\usepackage{textcomp}
\usepackage{stfloats}
\usepackage{url}
\usepackage{verbatim}
\usepackage{subcaption}
\usepackage{graphicx}
\usepackage{cite}
\usepackage{xcolor,soul,framed} 
\usepackage{multirow}
\usepackage{pgfplots}
\usepackage{xr}
\makeatletter

\newcommand*{\addFileDependency}[1]{
\typeout{(#1)}
\@addtofilelist{#1}
\IfFileExists{#1}{}{\typeout{No file #1.}}
}\makeatother

\pgfplotsset{compat=1.18}

\begin{document} 

\title{YOLO-Vehicle-Pro: A Cloud-Edge Collaborative Framework for Object Detection in Autonomous Driving under Adverse Weather Conditions}

\author{Xiguang Li , Jiafu Chen , Yunhe Sun , Na Lin , Ammar Hawbani , Liang Zhao

\thanks{ Xiguang Li, Jiafu Chen, Yunhe Sun, Ammar Hawbani, and Liang Zhao are with the School of Computer Science, Shenyang Aerospace University, Shenyang 110136, China (e-mail: lixiguang@sau.edu.cn, chenjiafu@stu.sau.edu.cn, sunyunhe@sau.edu.cn, anmande@ustc.edu.cn, lzhao@sau.edu.cn).
}
\thanks{Liang Zhao and Ammar Hawbani are the corresponding authors.}

\thanks{This paper has supplementary downloadable material available at http://ieeexplore.ieee.org, provided by the authors.}

}


\maketitle

\begin{abstract}
With the rapid advancement of autonomous driving technology, efficient and accurate object detection capabilities have become crucial factors in ensuring the safety and reliability of autonomous driving systems. However, in low-visibility environments such as hazy conditions, the performance of traditional object detection algorithms often degrades significantly, failing to meet the demands of autonomous driving. To address this challenge, this paper proposes two innovative deep learning models: YOLO-Vehicle and YOLO-Vehicle-Pro. YOLO-Vehicle is an object detection model tailored specifically for autonomous driving scenarios, employing multimodal fusion techniques to combine image and textual information for object detection. YOLO-Vehicle-Pro builds upon this foundation by introducing an improved image dehazing algorithm, enhancing detection performance in low-visibility environments. In addition to model innovation, this paper also designs and implements a cloud-edge collaborative object detection system, deploying models on edge devices and offloading partial computational tasks to the cloud in complex situations. Experimental results demonstrate that on the KITTI dataset, the YOLO-Vehicle-v1s model achieved 92.1\% accuracy while maintaining a detection speed of 226 FPS and an inference time of 12ms, meeting the real-time requirements of autonomous driving. When processing hazy images, the YOLO-Vehicle-Pro model achieved a high accuracy of 82.3\% mAP\textsubscript{@50} on the Foggy Cityscapes dataset while maintaining a detection speed of 43 FPS. 
\end{abstract}

\begin{IEEEkeywords}
Autonomous Driving, Object Detection, Cloud-Edge Collaboration, Intelligent Transportation Systems, Multi-Modal Feature Fusion, YOLO-Vehicle-Pro
\end{IEEEkeywords}

\section{INTRODUCTION}
\IEEEPARstart {O}{bject} Detection (OD) serves as a crucial component in Autonomous Driving (AD) and Intelligent Transportation Systems (ITS) \cite{1}, playing a vital role in ensuring system safety and reliability \cite{2}. The primary tasks of OD include rapid and accurate identification and localization of objects in images, as well as precise estimation of object bounding boxes and categories, providing reliable input for subsequent decision-making and planning. OD finds extensive applications in AD and ITS, such as obstacle avoidance \cite{3}, behavior prediction \cite{4}, and road condition assessment \cite{5}. However, OD still faces significant challenges in complex real-world scenarios, particularly under adverse weather conditions.

\begin{figure}[htbp]
\centering 
\begin{minipage}[t]{0.5\textwidth}
\begin{minipage}[t]{0.49\linewidth}
    \centering
    \begin{minipage}{\linewidth}
        \centerline{\textbf{Scenario 1}}
        \vspace{0.5em}  
        \centering
        \includegraphics[width=\linewidth,height=3cm,keepaspectratio]{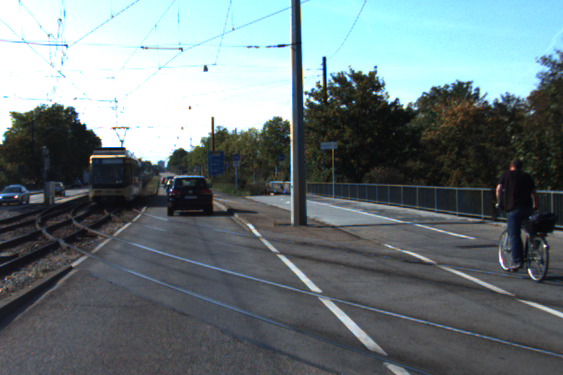}
        \centerline{(a) Scene a}
    \end{minipage}
\end{minipage}%
\hfill
\begin{minipage}[t]{0.49\linewidth}
    \centering
    \begin{minipage}{\linewidth}
        \centerline{\textbf{Scenario 2}}
        \vspace{0.5em}  
        \centering 
        \includegraphics[width=\linewidth,height=3cm,keepaspectratio]{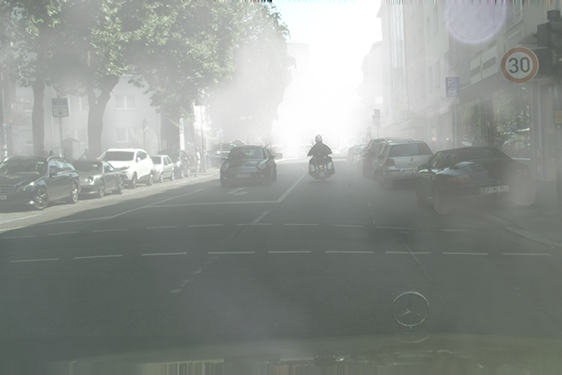}
        \centerline{(b) Scene b}
    \end{minipage}
\end{minipage}
\caption{Weather condition challenges in autonomous driving scenarios. (a) Clear weather environment (Scene a) and (b) hazy environment (Scene b), where low visibility in hazy conditions leads to reduced image contrast and blurred object contours.}
\label{fig:weather_conditions}
\end{minipage}
\end{figure}

Existing OD methods can be categorized into single-stage and two-stage approaches. Single-stage methods, such as YOLO \cite{6},\cite{7},\cite{8},\cite{9} and SSD \cite{10},\cite{11},\cite{12},\cite{13},\cite{40} integrate localization and classification tasks, achieving real-time detection capabilities more suitable for autonomous driving scenarios. Two-stage methods, centered around Region Proposal Networks (RPN), employ a ``propose-verify'' strategy, offering high accuracy but limited real-time performance. Nevertheless, these methods exhibit significant performance degradation in adverse conditions, such as Fig.\ref{fig:weather_conditions}.

To address this issue, researchers have recently proposed a series of innovative Image Dehazing Methods (IDM), which can also be categorized into two main classes: traditional methods based on prior knowledge and modern methods based on deep learning. Traditional methods based on prior knowledge, such as Dark Channel Prior (DCP) \cite{14}, Color Attenuation Prior (CAP) \cite{15}, and Non-local Color Prior (NCP) \cite{16}, rely on atmospheric scattering models and image prior information to estimate and remove haze. These methods have a solid theoretical foundation and high computational efficiency. However, they struggle to maintain stable performance when faced with complex and variable real-world environments, with a high probability of failure. In contrast, modern methods based on deep learning, such as All-in-One Dehazing Network (AOD-Net) \cite{17}, Multi-Scale Convolutional Neural Networks (MSCNN) \cite{18}, methods based on Generative Adversarial Networks (e.g., FD-GAN) \cite{19}, and feature fusion networks \cite{20}, \cite{42}, \cite{43}, \cite{44}, directly learn dehazing mapping relationships through end-to-end training. These methods possess powerful feature learning capabilities and adaptability to complex scenes but also face challenges such as high computational resource requirements and demanding training data needs. Therefore, how to effectively combine these dehazing methods with OD algorithms to improve the performance of AD systems under adverse weather conditions has become an important research direction.

Through summarizing and reflecting on the literature regarding the application of object detection in autonomous driving scenarios, this paper identifies that (1) accuracy is the primary focus of object detection, while real-time performance is largely overlooked, and (2) existing object detection algorithms perform well under normal weather conditions but experience significant performance degradation in adverse environments (such as haze, low light, etc.), making it challenging to meet the stability requirements of cloud-edge collaborative intelligent transportation perception systems.

In light of the above, aiming to address the challenges of object detection from clear to low-visibility environmental conditions, this paper proposes two innovative deep learning models: YOLO-Vehicle and YOLO-Vehicle-Pro, and achieves good results by deploying and testing them in conjunction with a self-built cloud-edge collaborative system. YOLO-Vehicle is an object detection model specifically optimized for autonomous driving scenarios, designed to provide efficient and accurate vehicle recognition capabilities. Building upon this foundation, this paper further improves the model structure and develops YOLO-Vehicle-Pro, an enhanced version specifically designed for hazy driving scenarios. YOLO-Vehicle-Pro significantly improves detection performance in low-visibility environments by introducing an improved image dehazing algorithm and adaptive feature extraction mechanism. The main contributions of this paper are as follows:

\begin{itemize}
\item We propose a novel object detector, YOLO-Vehicle model, comprising two main structures: an image processing module and a text processing module. The image processing module employs multi-scale feature map extraction techniques, effectively handling vehicle targets of different sizes and distances. The text processing module achieves effective utilization of textual information in the scene through region-text feature extraction, enhancing the model's understanding of complex traffic scenarios. 

\item We present YOLO-Vehicle-Pro, an advanced iteration of the YOLO-Vehicle model that provides significant enhancements to object detection systems specifically tailored for challenging hazy driving scenarios. The improved model contains two key innovations: an improved image dehazing algorithm and an adaptive feature extraction mechanism. These advances work synergistically to greatly improve detection performance in low-visibility environments.

\item We implemented a cloud-edge collaborative object detection system. In this system, YOLO-Vehicle model and YOLO-Vehicle-Pro model can be deployed on edge devices. The edge device is responsible for real-time image acquisition and preliminary processing, and when it encounters hazy weather, the system offloads part of the computing tasks to the cloud server. The cloud is mainly responsible for accelerating computationally intensive tasks performance.
\end{itemize}

The rest of this paper is organized as follows. Section \ref{sec_relatedwork} presents related work on object detection using deep learning networks and development work on edge computing devices implemented in vehicles. Section \ref{sec_system}, this paper introduces the YOLO-Vehicle and YOLO-Vehicle-Pro models. Section \ref{sec_experiment}, the paper presents the experimental results and analysis. In Section \ref{sec_conclusion}, the paper presents the conclusions of the paper.Note that
this work has a supplementary file, and the figures, tables,
and algorithms with underlined labels are cross-referenced
from the supplementary file.

\section{RELATED WORK}\label{sec_relatedwork}
This section introduces related work in object detection technologies and image dehazing. The related work involved in this study mainly includes two aspects: Object Detection (OD) methodologies and Image Dehazing Methods (IDM). Object detection techniques have evolved from traditional approaches such as SSD and Faster R-CNN to more advanced models like RetinaNet and DETR, which are designed to handle complex scenarios. This progression continues with the ongoing innovations in the YOLO series (YOLOV6, YOLOv7, YOLOV8) and extends to the latest multi-modal fusion approaches (e.g., ViLBERT and CLIP). Image dehazing technologies comprise prior-based methods (such as Dark Channel Prior (DCP) and color line-based approaches) and deep learning-based techniques (including AOD-Net, gated fusion networks, and semi-supervised methods). The related work has been moved to the supplementary file.

Compared to prior-based methods, deep learning approaches effectively dehaze images and restore object contours, ensuring reliable subsequent object detection. In this paper, we propose the YOLO-Vehicle-Pro model, which incorporates a deep learning-based dehazing module to better resolve conflicts between image distortion, blurring, and detail loss. Specifically, we introduce an improved feature fusion mechanism that retains more image details during the dehazing process. This mechanism combines multi-scale feature extraction with attention mechanisms, enabling the model to adaptively focus on crucial areas in the image, thereby enhancing object clarity and recognizability while dehazing. Additionally, we adopt a residual learning strategy, introducing residual connections to further alleviate the vanishing gradient problem in deep network training.

\section{SYSTEM PLATFORM AND ALGORITHM}\label{sec_system}
This section introduces the core technical details of YOLO-Vehicle. First, we present the independently developed edge-cloud collaborative autonomous vehicle object detection system architecture. Then, we provide an overview of the overall architectural approach. Subsections C and D respectively introduce the submodule design and loss function design of YOLO-Vehicle and YOLO-Vehicle-Pro.

\subsection{System Platform Design}
\begin{figure}[htbp]
\centering
\centerline{\includegraphics[width=0.5\textwidth]{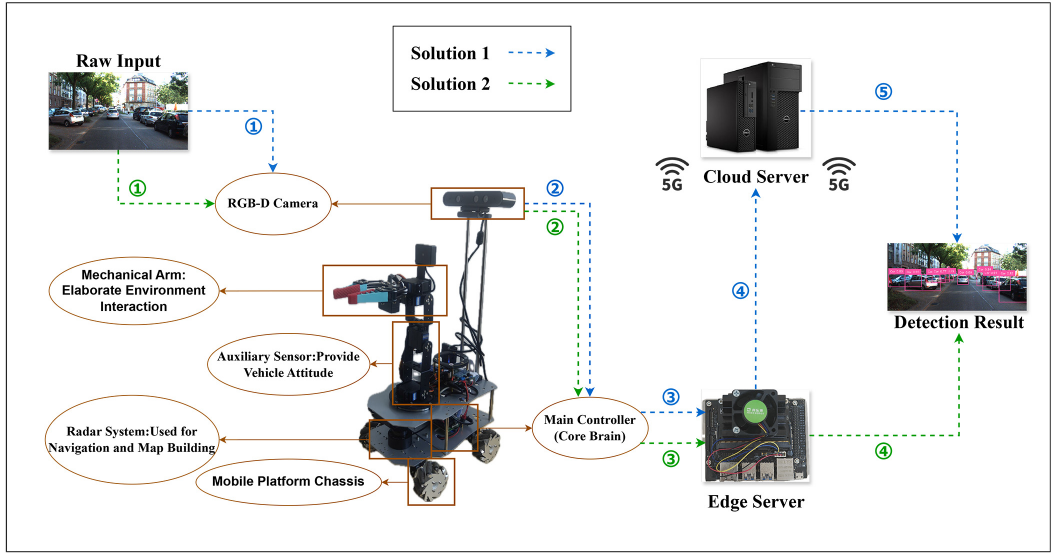}}
\caption{Edge-Cloud collaborative autonomous vehicle object detection system architecture}
\label{fig:car}
\end{figure}

This paper proposes an innovative distributed computing architecture aimed at optimizing real-time object detection systems in autonomous vehicles. As shown in the Fig.\ref{fig:car}, this architecture combines the advantages of edge computing and cloud computing, forming an efficient and flexible intelligent transportation solution. The core of the system is constituted by a mobile platform that integrates various sensors and computing units. Among these, the RGB-D camera serves as the primary visual input device, responsible for capturing rich visual information of the environment. Auxiliary sensors provide vehicle posture data, while the radar system is used for navigation and map construction. These components collectively ensure comprehensive environmental perception for the vehicle. Notably, this system adopts a two-tier computing architecture. The edge server, acting as the on-board main controller, bears the responsibility of real-time data processing and decision-making. This design significantly reduces data transmission latency and improves system response speed. Simultaneously, the cloud server maintains connection with edge devices via a 5G network, handling more complex computational tasks such as model training and optimization.

A key advantage of this layered architecture lies in its ability to balance computational demands and energy efficiency. Edge devices focus on low-latency inference tasks, while high-intensity training work is transferred to the cloud. This not only optimizes the energy consumption of the on-board system but also ensures continuous model updates and performance improvements. The two solution paths shown in the Fig.\ref{fig:car} demonstrate the system's flexibility. Solution 1 represents a processing flow more reliant on cloud computing, while Solution 2 utilizes edge computing capabilities to a greater extent. This flexibility allows the system to dynamically adjust its working mode according to different application scenarios and network conditions.

Fig.\ref{fig:car} presents a novel autonomous vehicle real-time object detection system architecture, integrating edge computing and cloud computing paradigms. The Fig. \ref{fig:car} details has been moved to the supplementary file. 

\subsection{Overall Framework}
\begin{figure*}[htbp]
\centerline{\includegraphics[width=\linewidth]{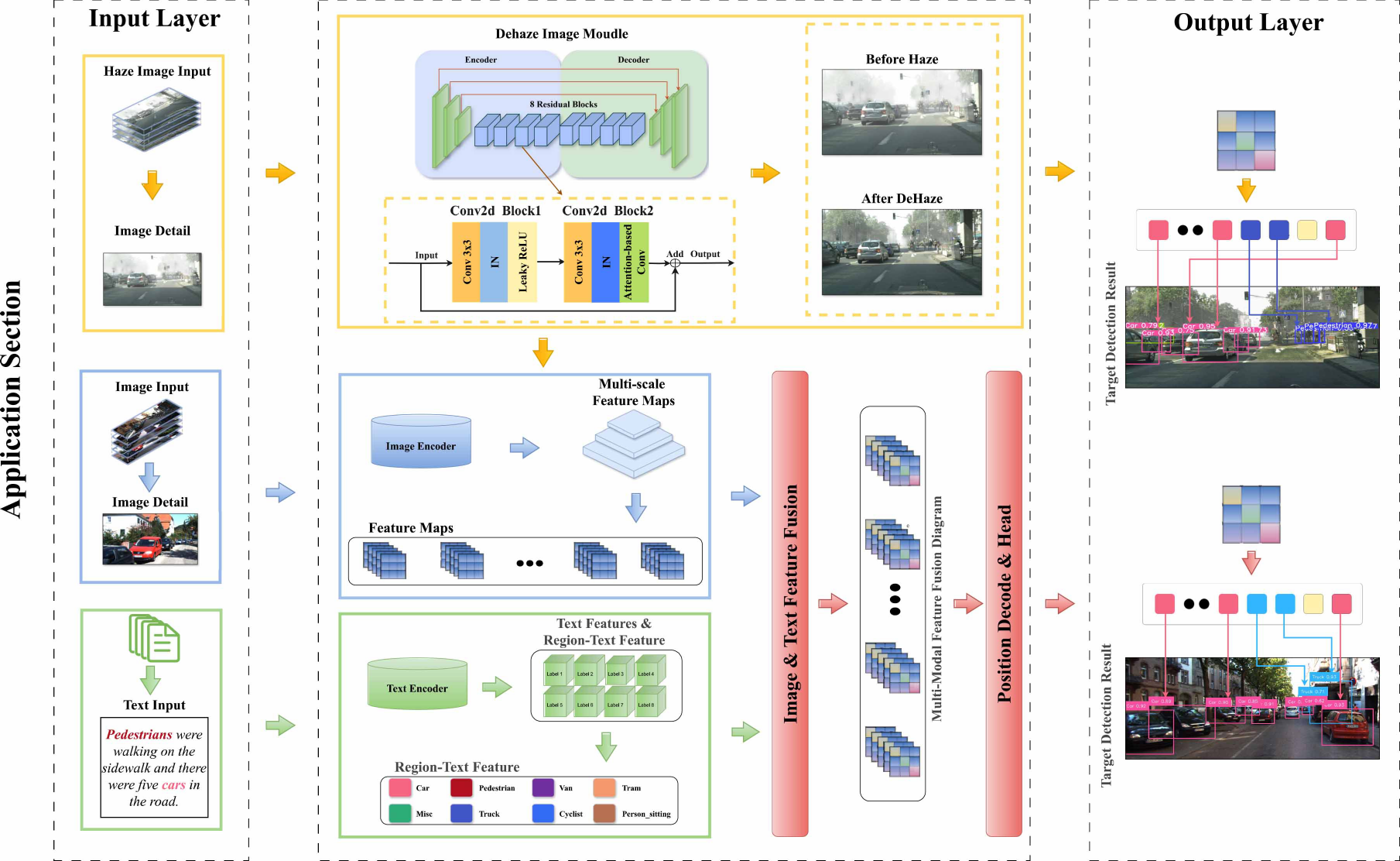}}
\caption{Structure and workflow of YOLO-Vehicle and YOLO-Vehicle-Pro models}
\label{fig3}
\end{figure*}

Prior to the object detection task, traditional object detection methods (e.g., YOLOV6\cite{6}, YOLOV7\cite{7}, YOLOV8\cite{24}) perform image feature extraction through a backbone network. While this approach excels in ideal image input scenarios, its performance is significantly affected in complex traffic road scenes, especially under hazy weather conditions. To overcome this limitation, this paper proposes an innovative multi-modal fusion framework, as shown in Fig.\ref{fig3}, combining image processing, text processing, and efficient dehazing techniques.

Drawing inspiration from multi-modal fusion ideas \cite{9},\cite{46}, this paper proposes an innovative application framework that introduces diverse data sources at the input layer. It includes three key components: text input, standard image input, and hazy image input. The Fig. \ref{fig3} details has been moved to the supplementary file.

\subsection{YOLO-Vehicle Network Structure}
YOLO-Vehicle is an object detection model based on the fusion of text and image information. The overall architecture of this model comprises four main components: a text feature extractor, an image feature extractor, a multi-modal feature fusion module, and a detection head. The detailed content of the model architecture is as follows:

\subsubsection{Text Feature Extraction}
The text feature extractor in the YOLO-Vehicle model uses Clip-VIT, a pre-trained model based on the Vision Transformer (ViT) architecture \cite{31}. This model utilizes self-attention mechanisms and multi-head attention layers to process input text, achieving efficient text-image contrastive learning. During processing, the input text first undergoes tokenization and is then embedded into a high-dimensional vector space. Subsequently, these embedded vectors pass through a series of Transformer encoder layers, each containing self-attention sub-layers and feed-forward neural network sub-layers. Finally, the model outputs a fixed-dimension text feature vector \(F_{text}\) with a dimension of 512. This feature vector captures the semantic information of the input text and can be represented as in \eqref{eq_aa}, where \(E_t(x)\) represents the word embedding function for the text, and \(x\) is the input text.

\begin{equation} \label{eq_aa}
F_{text} = \text{CLIP-ViT}(E_t(x)) \in \mathbb{R}^{1 \times 512}
\end{equation}

 The core self-attention mechanism of CLIP-ViT is formalized as in \eqref{eq_2}, where \(Q\), \(K\), \(V\) represent the query, key, and value matrices respectively, and \(d_k\) is the dimension of the key vector.

\begin{equation} \label{eq_2}
\text{Attention}(Q,K,V) = \text{softmax}\left(\frac{QK^T}{\sqrt{d_k}}\right)V
\end{equation}

To further enhance the performance of the CLIP-ViT-base-patch32 model in the transportation domain, this paper conducted targeted fine-tuning. A large-scale dataset containing hundreds of thousands of traffic-related descriptions was constructed. These descriptions cover multiple aspects such as vehicle type, color, brand, model, and features, for example, "red sedan", "large freight truck", "blue electric bicycle", etc. The fine-tuning adopted a contrastive learning strategy, pairing traffic semantic prompts with corresponding vehicle images to form positive samples, while randomly selecting mismatched text-image pairs as negative samples. The optimization objective is to maximize the similarity of positive sample pairs while minimizing the similarity of negative sample pairs. This is formally represented as in \eqref{eq_3}, where \(s(t,i)\) is the similarity score between text \(t\) and image \(i\), \(i_+\) is the matching image, \(i\) is the mismatched image, and \(\tau\) is the temperature parameter.

\begin{equation} \label{eq_3}
\mathcal{L}_{\text{contrastive}} = -\log \frac{\exp(s(t,i_+)/\tau)}{\sum_i \exp(s(t,i)/\tau)}
\end{equation}

Empirical analysis shows that the fine-tuned model improved accuracy by 15.3\% on vehicle-related text-image matching tasks and increased the F1 score by 12.7\% on vehicle attribute prediction tasks. These improvements led to a significant enhancement in the overall performance of the YOLO-Vehicle.

\subsubsection{Image Feature Extraction}
YOLO-Vehicle adopts YOLOV8 \cite{24} as its image feature extractor, utilizing its advanced CSPDarknet53 backbone network and feature pyramid structure to extract multi-scale visual features. This model integrates spatial and channel attention mechanisms \cite{45}, enhancing feature expression capabilities. YOLOV8 generates three different scale feature maps \(F_{img} = \{f_1, f_2, f_3\}\), where \(f_i \in \mathbb{R}^{C \times H_i \times W_i}\), for detecting vehicle targets of different sizes. It leverages weights pre-trained on large-scale object detection datasets (such as COCO) and then fine-tunes on specialized vehicle datasets to adapt to specific vehicle detection tasks.

\subsubsection{Multi-modal Feature Fusion}
A key innovation point of the YOLO-Vehicle model lies in its efficient multi-modal feature fusion strategy. To achieve effective integration of heterogeneous modal features, cross-modal alignment techniques are employed, mapping features from different modalities to a unified semantic embedding space through linear projection. To optimize the model's computational efficiency and reduce parameter complexity, extensive experiments were conducted. The optimal configuration balancing expressive power and computational overhead was ultimately determined, setting the output tensor shape of the text feature encoder to $(1, 512)$. The feature alignment expressions as in \eqref{eq_4}, \eqref{eq_5}, where $W_{img}$, $W_{text}$ are learnable weight matrices, and $b_{img}$, $b_{text}$ are bias terms.

\begin{equation} \label{eq_4}
{F^\prime}_{img}=W_{img} \cdot F_{img}+b_{img} \in \mathbb{R}^{1\times512}
\end{equation}
\begin{equation} \label{eq_5}
{F^\prime}_{text}=W_{text} \cdot F_{text}+b_{text} \in \mathbb{R}^{1\times512}
\end{equation}

After obtaining the aligned feature information, cross-modal correlation is calculated using equation \eqref{eq_2}, where $Q$ comes from image features, $K$ and $V$ come from text features, $\sigma$ is the sigmoid activation function, and $\otimes$ represents element-wise multiplication. The $F_{fused}$ are calculated by \eqref{eq_6}.

\begin{equation} \label{eq_6}
\begin{aligned}
F_{fused} = \sigma(&W[F'_{img}; F'_{text}] + b) \otimes F'_{img} \\
           &+ (1-\sigma) \otimes \text{Attention}(F'_{img}, F'_{text}, F'_{text})
\end{aligned}
\end{equation}

\subsubsection{Position Encoding and Detection Head}
After multi-modal feature fusion, the YOLO-Vehicle model introduces an efficient position encoding mechanism, adopting a fixed encoding scheme constructed with sine and cosine functions \cite{41}. Specifically, for the fused feature $F_{fused}\in \mathbb{R}^{1\times512}$, each element of the position encoding $PE\in \mathbb{R}^{1\times512}$ is defined by \eqref{eq_7}, \eqref{eq_8}, where $pos$ represents the position index on the feature map, and $i$ is the dimension index. 

\begin{equation} \label{eq_7}
{PE}_{(pos,2i)}=\sin{(pos/{10000}^{2i/512})}
\end{equation}
\begin{equation} \label{eq_8}
{PE}_{(pos,2i+1)}=\cos{(pos/{10000}^{2i/512})}
\end{equation}

The advantage of this encoding method is that it can generate a unique representation for each position while maintaining the linear relationship of relative positions. Finally, $F_{final}$ are calculated by \eqref{eq_9}.

\begin{equation} \label{eq_9}
F_{final}=F_{fused}+PE
\end{equation}

The model effectively injects spatial information into the fused features. This not only allows subsequent attention mechanisms to utilize positional correlations but also maintains the translation equivariance of the features. After cross-modal attention and fusion operations, the final fused features need to be reconstructed to the shape of the original image features to be compatible with the detection head as in \eqref{eq_10}.

\begin{equation} \label{eq_10}
F_{output}=\text{Reshape}(F_{final},(C,H,W))
\end{equation}

The final output, denoted as \( F_{final} \), represents the fused feature, which is reshaped into a tensor of dimensions \( C \times H \times W \) through a reshape operation. This reshaped feature, referred to as \( F_{output} \), which is subsequently forwarded to the detection head for the object detection task. The detection head produces three key components: object prediction (\( P_{obj} \)), bounding box regression (\( B \)), and class prediction (\( C \)), as defined by the equations \eqref{eq_11}, \eqref{eq_12}, and \eqref{eq_13}.

\begin{equation} \label{eq_11}
P_{obj}=\sigma\left(W_{obj}\ast F_{output}+b_{obj}\right)
\end{equation}

\begin{equation} \label{eq_12}
B=W_{box}\ast F_{output}+b_{box}
\end{equation}

\begin{equation} \label{eq_13}
C=\text{softmax}(W_{cls}\ast F_{output})+b_{cls}
\end{equation}

\subsection{YOLO-Vehicle-Pro Network Structure}
\begin{figure}
\centering
\includegraphics[width=0.5\textwidth]{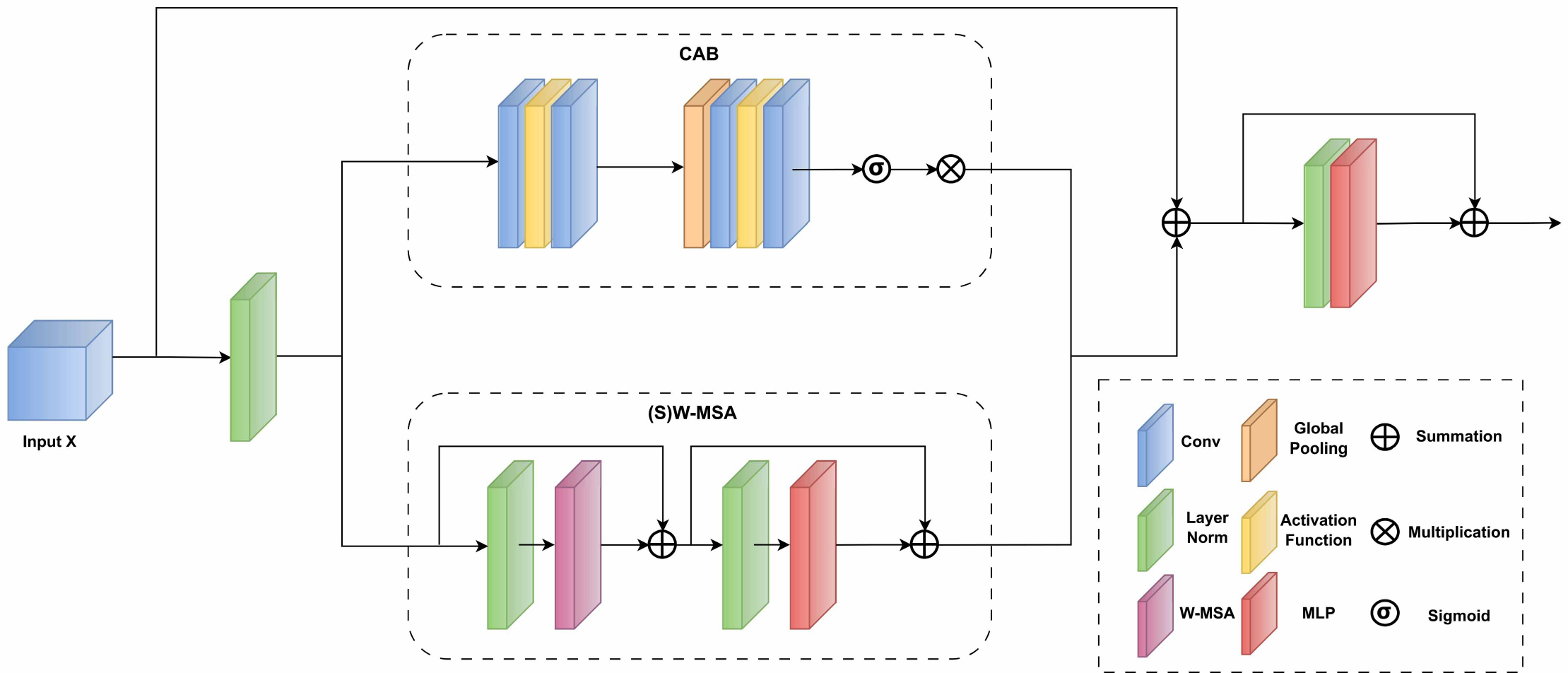}
\caption{Architecture of Attention-based Conv}
\label{fig:attention_conv}
\end{figure}

In the YOLO-Vehicle-Pro network, to enhance the model's performance under adverse weather conditions, particularly in hazy environments, an efficient dehazing module is introduced prior to image feature extraction. This module aims to restore hazy images to a quality close to that of normal weather conditions, thereby providing clearer input for subsequent feature extraction and object detection. Fig.\ref{fig:attention_conv} details the core design of the dehazing module.

\subsubsection{Attention of Attention-based Convolution}

The input feature $X$ first passes through a shared feature extraction layer, then is fed into two parallel branches: CAB and (S)W-MSA \cite{32}. The CAB branch employs a multi-layer convolution structure to capture inter-channel dependencies by gradually extracting and refining features. Its output passes through a Sigmoid activation function, generating channel attention weights. These weights are then element-wise multiplied with the original input, achieving adaptive feature enhancement in the channel dimension.

The (S)W-MSA branch focuses on capturing contextual information in the spatial dimension. It first processes the input through two parallel feature transformation modules, then adds the results. This branch adopts a windowed multi-head self-attention mechanism ((S)W-MSA), which efficiently models long-range dependencies within local regions. By dividing the input into multiple windows and applying self-attention within each window, it captures rich spatial contextual information while maintaining computational efficiency.

The outputs from both branches are fused through element-wise addition, combining attention information from both channel and spatial dimensions. Finally, the fused features pass through a convolution layer for final feature adjustment and integration. This design allows the model to flexibly focus on important image regions and features at different scales and dimensions, thereby more effectively handling complex dehazing tasks.

\subsubsection{Activation Function}
Another improvement in the dehazing module implementation is the replacement of the traditional ReLU activation function \cite{33} with Leaky ReLU \cite{34}. Through extensive experimental validation, this paper demonstrates that the introduction of Leaky ReLU addresses the "dying ReLU" problem faced by ReLU. Unlike ReLU, which sets activation to zero for all negative values, Leaky ReLU introduces a small slope (typically 0.01) in the negative range. Its mathematical expression can be represented as in \eqref{eq_14}. Where $\alpha$ is a small positive constant.

\begin{equation} \label{eq_14}
f(x)=\begin{cases}
x, & x > 0 \\
\alpha x, & x \leq 0
\end{cases}
\end{equation}

In the specific implementation of the attention convolution module, Leaky ReLU is applied after the convolution layers in both the CAB (Channel Attention Branch) and (S)W-MSA ((Shifted) Window Multi-head Self-Attention) branches. This ensures that more information flow is preserved during the process of feature extraction and attention weight calculation, thereby enhancing the model's expressive power and learning efficiency.

Furthermore, Leaky ReLU influences the model's initialization strategy and learning rate selection. Due to its non-zero gradient in the negative range, this paper adjusted the weight initialization method to ensure that the network can fully utilize the characteristics of Leaky ReLU during the initial stages of training.

\subsection{Loss Functions}

\subsubsection{YOLO-Vehicle Loss Function}
The loss function of YOLO-Vehicle comprises three components: classification loss, bounding box regression loss, and distribution focal loss. Specifically, the loss function is represented as in \eqref{eq_15}. Where $\lambda_1$, $\lambda_2$ and $\lambda_3$ are weight coefficients used to balance the contributions of different loss terms. $L_{cls}$ is the classification loss, $L_{bbox}$ is the bounding box regression loss, and $L_{dfl}$ is the distribution focal loss.

\begin{equation} \label{eq_15}
L_{detect}={\lambda_1\cdot L}_{cls}+\lambda_2\cdot L_{bbox}+\lambda_3{\cdot L}_{dfl}
\end{equation}

The classification loss evaluates the model's prediction accuracy for target categories. In object detection, this typically involves predicting whether each candidate box contains an object (foreground vs. background) and specifically which class of object it is. The bounding box regression loss assesses the difference between the predicted bounding boxes and the ground truth boxes. The distribution focal loss is a relatively new loss function primarily used to improve the accuracy of bounding box regression.

In this paper, the classification loss adopts cross-entropy loss with sigmoid activation, which is suitable for multi-label classification problems, allowing a sample to belong to multiple categories simultaneously. The bounding box loss employs CIoU loss \cite{35}, which, compared to ordinary IoU loss \cite{36}, can better handle objects of different scales and shapes. The distribution focal loss transforms the bounding box regression problem into a probability distribution prediction problem, enabling better handling of uncertainties in bounding box predictions. This contributes to improved performance in small object detection and overall accuracy of bounding box regression.

In the experiments, the weight coefficients were set to $\lambda_1=0.6$, $\lambda_2=7$, and $\lambda_3=0.4$.

\subsubsection{YOLO-Vehicle-Pro Dehazing Module Loss Function}
The loss function definition for the dehazing module in YOLO-Vehicle-Pro includes four losses: adversarial loss, patch-based contrastive loss, self-contrastive perceptual loss, and identity loss. Specifically, the loss function is represented as in \eqref{eq_16}. Where $\lambda_4$, $\lambda_5$, $\lambda_6$, and $\lambda_7$ are weight coefficients. $L_{adv}(G)$ is the adversarial loss, using vanilla GAN loss to train the generator and discriminator, promoting more realistic dehazed images. $L_{pc}$ is the patch-based contrastive loss, which maximizes the mutual information between corresponding patches of input and output images. $L_{SCP}$ is the self-contrastive perceptual loss, used to encourage the restored image to be closer to the real clear image and farther from the hazy image. $L_{ide}$ is the identity loss, used to maintain consistency between the structure of the dehazed image and the input.

\begin{equation} \label{eq_16}
L_{dehaze}={\lambda_4\cdot L}_{adv}(G)+\lambda_5\cdot L_{pc}+\lambda_6{\cdot L}_{SCP}+\lambda_7{\cdot L}_{ide}
\end{equation}

This combination of loss functions allows the network to learn to remove haze from real-world hazy images in an unsupervised manner while maintaining image naturalness and details. The introduction of contrastive learning enables the network to utilize unpaired real-world clear and hazy images for training, thereby improving the model's generalization capability in real-world scenarios. Each loss function plays a crucial role in the overall training process, collectively contributing to the enhancement of model performance.

\section{EXPERIMENT AND ANALYSIS}\label{sec_experiment}

This section begins by introducing the process of constructing an autonomous driving image dataset and the evaluation metrics for the experimental results. Subsequently, through extensive experiments, this paper compares the proposed lightweight object detection network (YOLO-Vehicle) with existing mainstream lightweight object detection networks (namely Edge YOLO \cite{1}, YOLOV6 \cite{6}, YOLOV8 \cite{24} and YOLO-World \cite{9})\footnote{ In this paper,our newly trained model and other comparison models are available on https://github.com/chenjiafu-George/YOLO-Vehcile-Pro}.

\subsection{Experimental Datasets and Hardware Platform}

This paper introduces three datasets, namely COCO2017, KITTI, and Foggy Cityscapes dataset. The specific description content has been moved to the supplementary file.

Table \ref{tab:datasets} provides a comprehensive overview of the three datasets used in this paper, detailing the following characteristics for each dataset: training dataset, test dataset configuration, annotation type, resolution size, image source, and dataset size.

\begin{table}[htbp]
\centering
\caption{Detailed Description of COCO2017, KITTI2D, and Foggy Cityscapes Datasets}
\label{tab:datasets}
\renewcommand{\arraystretch}{1.2}
\begin{tabular}{>{\centering\arraybackslash}p{2cm}|>{\centering\arraybackslash}p{1.8cm}>{\centering\arraybackslash}p{1.8cm}>{\centering\arraybackslash}p{1.5cm}}
\hline
\textbf{Datasets} & \textbf{COCO2017} & \textbf{KITTI} & \textbf{Foggy Cityscapes} \\
\hline
Classes& 80 & 8 & 8 \\
\hline
Training Dataset & 118,287 & 7,481 & 1,000 \\
\hline
Test Dataset & 5,000 & 7,518 & 500 \\
\hline
Annotation Type & Object Detection & Object Detection & Image Dehaze \\
\hline
Resolution Size & 640x640 & 1280x384 & 2048x1024 \\
\hline
Image Source & General Object Detection & Vehicle Vision, Auto Driving & Auto Driving \\
\hline
Dataset Size & 25GB & 12GB & 3GB \\
\hline
\end{tabular}
\end{table}

In the process of model training, experiments were conducted under the Ubuntu 18.04 operating system. For detailed system specifications, please refer to the Supplementary file. Table \ref{tab:Hardware_table} provides a comprehensive overview of the hardware platform configurations, including CPU, CPU core, GPU, GPU memory, GFLOPS (FP16), and accelerator libraries for four distinct platforms: training platform, autonomous driving platform, K8s cloud node platform, and K8s edge node platform. This information facilitates understanding of the performance and applicability of different platforms in model training and inference tasks.

\begin{table*}[]
\centering
\caption{Hardware Platform Description}
\renewcommand{\arraystretch}{1.5}
\begin{tabular}{c|ccccccc}
\hline
\textbf{Hardware Platform} & \textbf{Training Platform}   & \textbf{Autonomous Driving Platform} & \textbf{K8s Cloud Node Platform} & \textbf{K8s Edge Node Platform}  \\ \hline
CPU                  & Intel Core i9-10900K   & Quad-Core ARM Cortex-A57       & Intel Xeon E5-2630 v4      & Intel Xeon E5-2630 v4 \\ \hline
CPU Core             & 10                     & 4                              & 10                         & 1                     \\ \hline
GPU                  & NVIDIA GeForce GTX 3090 & NVIDIA Maxwell GPU             & NVIDIA GeForce GTX TITAN   & ×                     \\ \hline
GPU Memory           & 24 GB GDDR6X           & 4 GB LPDDR4                    & 6 GB GDDR5                 & ×                     \\ \hline
GFLOPS (FP16)        & 35,580 GFLOPS          & 472 GFLOPS                     & 4,500 GFLOPS               & ×                     \\ \hline
Accelerator Library  & CUDA 11.1, CUDNN 8     & CUDA 10.2, CUDNN 7.6           & CUDA 10.2, CUDNN 7.6       & ×                     \\ \hline
\end{tabular}
\label{tab:Hardware_table}
\end{table*}

The experimental results and analysis section systematically evaluates and compares the YOLO-Vehicle and YOLO-Vehicle-Pro models in the aforementioned computing environments. For each environment, the following performance metrics are examined: mean Average Precision (mAP), Frames Per Second (FPS), inference time, and resource utilization (including CPU usage, GPU usage, and memory usage). Through comprehensive analysis of these indicators, this paper provides a thorough evaluation of model performance under various computing conditions, offering a foundation for informed deployment decisions.

\subsection{Evaluation Metrics}

In this object detection task, a comprehensive suite of metrics is employed to assess model performance. These metrics include Precision, Recall, Average Precision (AP), mean Average Precision at 50\% Intersection over Union (mAP\textsubscript{@50}), mAP\textsubscript{@75}, Frames Per Second (FPS), Inference Time, and Model Size. Precision quantifies the accuracy of the model's positive predictions, indicating the proportion of correctly identified positive samples. Recall measures the model's ability to detect all positive instances, revealing its comprehensiveness in identification.

AP encapsulates the model's Precision and Recall across various threshold settings, essentially capturing the area beneath the precision-recall curve. True Positives (TP) denote correctly classified positive instances, while False Positives (FP) represent misclassified negative instances as positive. Conversely, False Negatives (FN) refer to positive instances overlooked by the model, and True Negatives (TN) are negative instances accurately recognized by the model.

mAP\textsubscript{@50} denotes the aggregate average precision for all classes at an Intersection over Union (IoU) threshold of 0.50, where IoU is the ratio of overlap between the predicted and actual bounding boxes. mAP\textsubscript{@75} extends this metric to an IoU threshold of 0.75, providing a more stringent measure of class-wise performance. FPS indicates the model's processing capability, measured by the number of frames processed per second. Inference Time, a critical metric for real-time applications, reflects the latency experienced by the model in analyzing a single image.

Model Size, typically expressed in megabytes, is an indicator of the model's complexity and its consequent resource demands. Collectively, Precision, Recall, AP, mAP\textsubscript{@50}, and mAP\textsubscript{@75} evaluate the model's accuracy, FPS and Inference Time gauge its efficiency, and Model Size assesses its resource consumption.

\subsection{Experimental Results and Analysis}

\begin{figure*}[htbp]
\centering 
\begin{minipage}[t]{1\textwidth}
\begin{minipage}[t]{0.5\linewidth}
    \begin{minipage}{\linewidth}
        \centering
        \includegraphics[width=\linewidth,height=7cm]{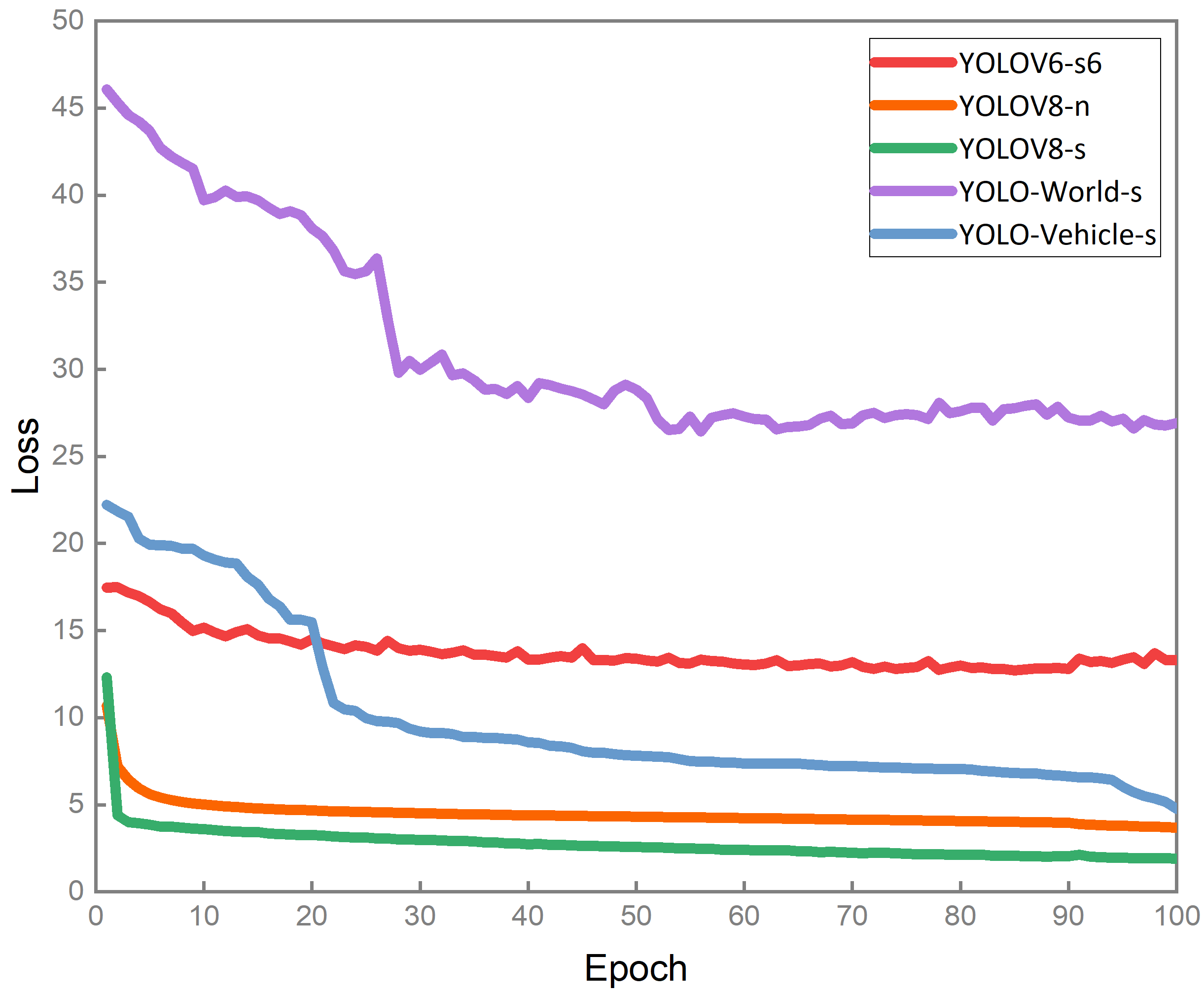}
        \caption{Training loss values under COCO dataset}
        \label{fig:coco_loss}
    \end{minipage}
\end{minipage}%
\hfill
\begin{minipage}[t]{0.5\linewidth}
    \begin{minipage}{\linewidth}
        \centering 
        \includegraphics[width=\linewidth,height=7cm]{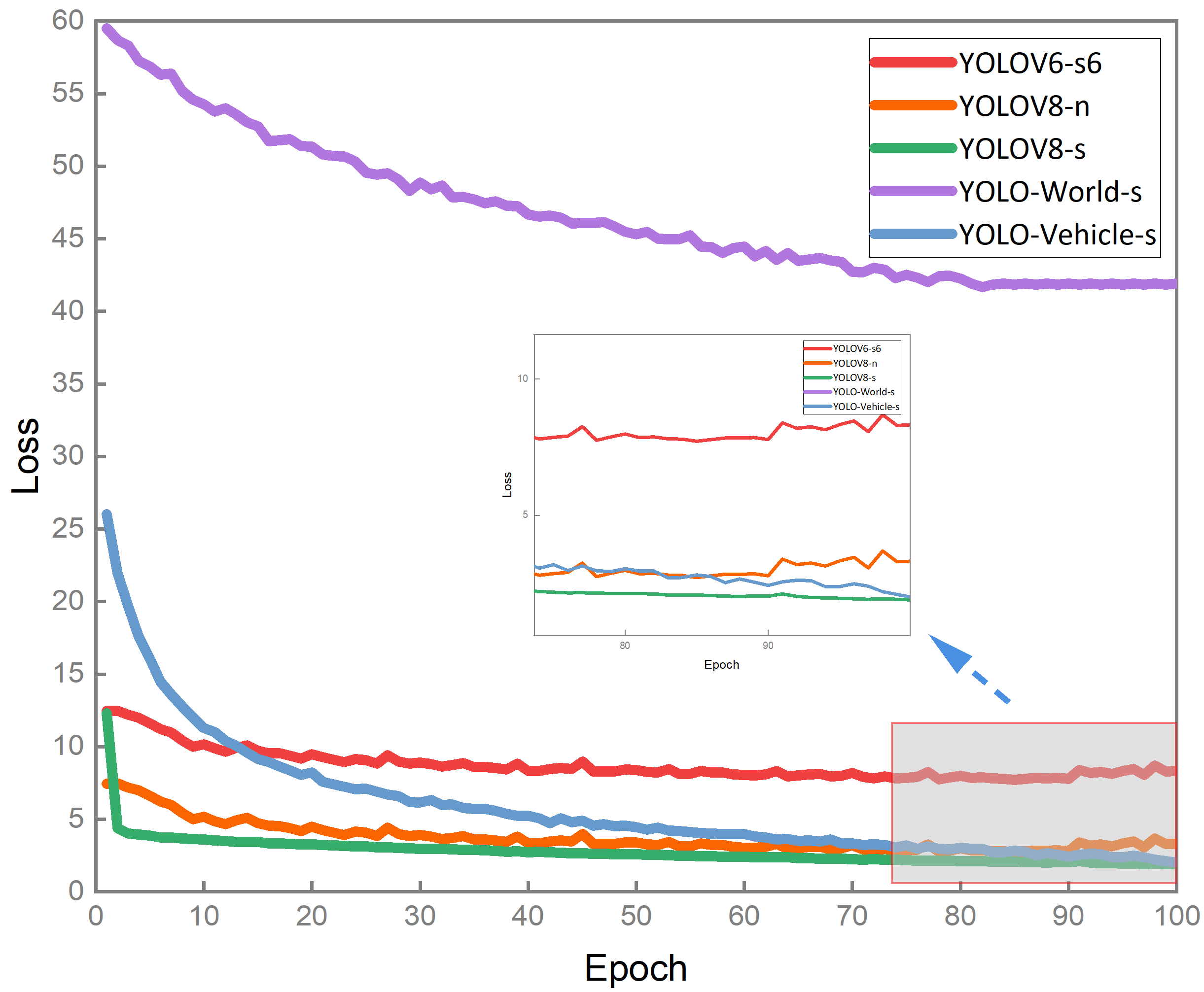}
        \caption{Training loss values under KITTI dataset}
        \label{fig:kitti_loss}
    \end{minipage}
\end{minipage}
\end{minipage}
\label{fig:combined_loss}
\end{figure*}

In this paper, the COCO2017 and KITTI dataset in Table \ref{tab:datasets} are selected for model training, which are widely recognized in the field of object detection. During the training process of the COCO2017 dataset, the YOLO-Vehicle model is carefully tuned for parameters. The batch size is set to 32 to strike a balance between training efficiency and memory usage. The weights of the loss function are set to reflect the importance of different tasks, where the weight of bounding box regression loss (loss\_bbox) is 7.5, the weight of classification loss (loss\_cls) is 0.5, and the weight of distribution focal loss (loss\_dfl) is 0.375. This weight configuration emphasizes the importance of boundary box localization while taking into account classification accuracy. The learning rate strategy uses an initial value of 0.001 and a learning rate factor of 0.01 to ensure the stable convergence of the model and avoid falling into local optimum. In addition, this paper introduces Mixup data augmentation technology with probability set to 0.15 to improve the generalization ability of the model by creating new mixed samples, which performs well especially when dealing with complex object detection tasks. Fig.~\ref{fig:coco_loss} shows the LOSS function values of YOLOV6, YOLOV8n, YOLOV8s, YOLO-World, and YOLO-Vehicle models trained under the COCO dataset.

\begin{figure}[htbp]
\centering 
\begin{minipage}[t]{0.5\textwidth}
\begin{minipage}[t]{0.5\linewidth}
    \begin{minipage}{\linewidth}
        \centering
        \includegraphics[width=\linewidth,height=3cm,keepaspectratio]{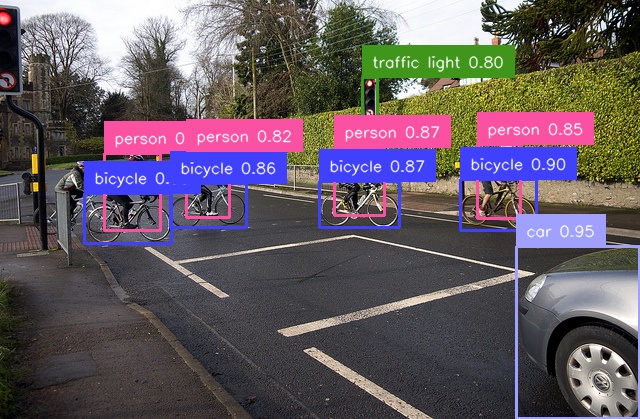}
        \centerline{(a) COCO Test 1}
        \label{fig:coco_1}
    \end{minipage}
    \begin{minipage}{\linewidth}
        \centering
        \includegraphics[width=\linewidth,height=3cm,keepaspectratio]{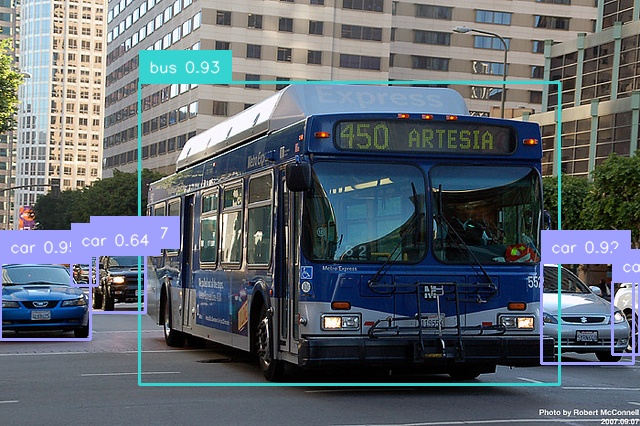}
        \centerline{(b) COCO Test 2}
        \label{fig:coco_2}
    \end{minipage}
\end{minipage}%
\hfill
\begin{minipage}[t]{0.5\linewidth}
    \begin{minipage}{\linewidth}
        \centering
        \includegraphics[width=\linewidth,height=3cm,keepaspectratio]{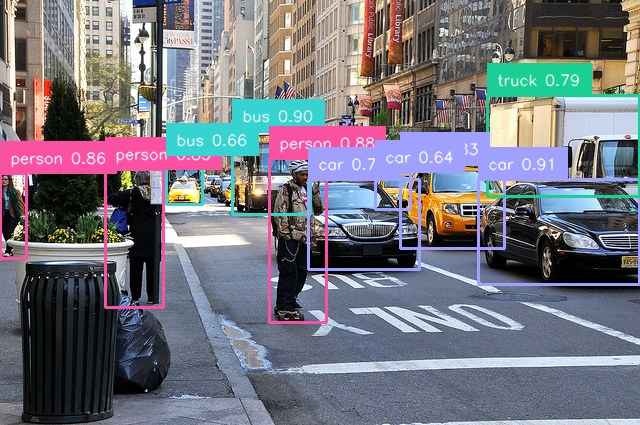}
        \centerline{(c) COCO Test 3}
        \label{fig:coco_3}
    \end{minipage}
    \begin{minipage}{\linewidth}
        \centering
        \includegraphics[width=\linewidth,height=3cm,keepaspectratio]{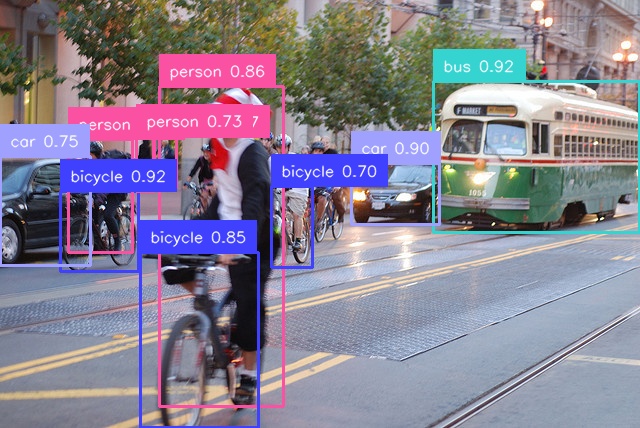}
        \centerline{(d) COCO Test 4}
        \label{fig:coco_4}
    \end{minipage}
\end{minipage}
\end{minipage}
\caption{YOLO-Vehicle-v1s detects part of the test examples in the COCO dataset}
\label{fig:coco}
\end{figure}

\begin{table*}[]
\centering
\caption{Comparison of the proposed method and the previous object detection networks on the COCO2017 dataset}
\renewcommand{\arraystretch}{1.5}
\begin{tabular}{c|ccccccc}
\hline
\textbf{Model}            & \textbf{Input    Size} & \textbf{mAP}  & \textbf{mAP\textsubscript{@50}} & \textbf{mAP\textsubscript{@75}} & \textbf{FPS} & \textbf{Inference    Time} & \textbf{Size(MB)} \\ \hline
YOLOV6-s6        & 640x640       & 43.9 & 60.3   & 47.2   & 294 & 11.3ms            & 87.3     \\ \hline
YOLOV8n          & 640x640       & 43.7 & 60.8   & 47.0   & 442 & 7ms               & 6.3      \\ \hline
YOLOV8s          & 640x640       & 44.4 & 61.2   & 48.1   & 386 & 8ms               & 21.5     \\ \hline
Edge YOLO{[}1{]} &               &      & 47.3   &        &     &                   &          \\ \hline
YOLO-World       & 640x640       & 44.8 & 61.6   & 48.5   & 183 & 28                & 523      \\ \hline
\textbf{YOLO-Vehicle-v1s} & 640x640       & \textbf{45.8} & \textbf{62.5}   & \textbf{49.7}   & \textbf{346} & \textbf{10ms}              & \textbf{43.2}     \\ \hline
\end{tabular}
\label{tab:coco_table}
\end{table*}

The results presented in Table~\ref{tab:coco_table} demonstrate the superior performance of the YOLO-Vehicle-v1s model across several key indicators. In terms of detection accuracy, the model achieves mAP of 45.8\%, with mAP\textsubscript{@50} and mAP\textsubscript{@75} reaching 62.5\% and 49.7\%, respectively, surpassing other models in the comparison. This indicates that YOLO-Vehicle-v1s maintains high detection accuracy across different IoU thresholds. Regarding processing speed, the model attains 346 FPS, second only to YOLOV8n, but significantly outperforming other models such as YOLO-World. The inference time for a single frame is 10ms, comparable to YOLOV8s and substantially better than YOLO-World's 28ms. Notably, YOLO-Vehicle-v1s achieves model lightweight while maintaining high performance. Its model size is 43.2MB, significantly smaller than YOLOV6-s6's 87.3MB and YOLO-World's 523MB. This compact design not only reduces storage requirements but also facilitates deployment on resource-constrained edge devices.

\begin{figure*}[htbp]
\centering 
\begin{minipage}[t]{1\textwidth}
\begin{minipage}[t]{0.33\linewidth}
    \begin{minipage}{\linewidth}
        \centering
        \includegraphics[width=\linewidth,height=4cm,keepaspectratio]{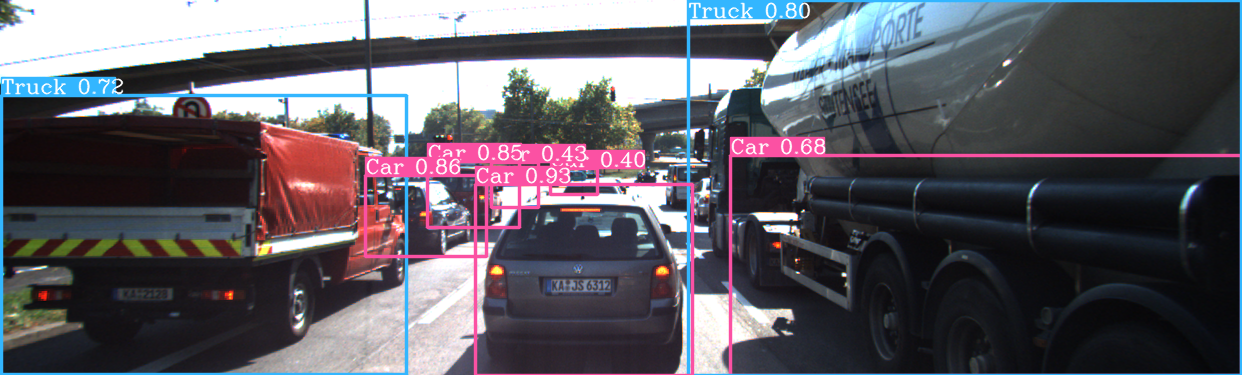}
        \centerline{(a) 1.YOLOV6-s6}
    \end{minipage}

    \begin{minipage}{\linewidth}
        \centering
        \includegraphics[width=\linewidth,height=4cm,keepaspectratio]{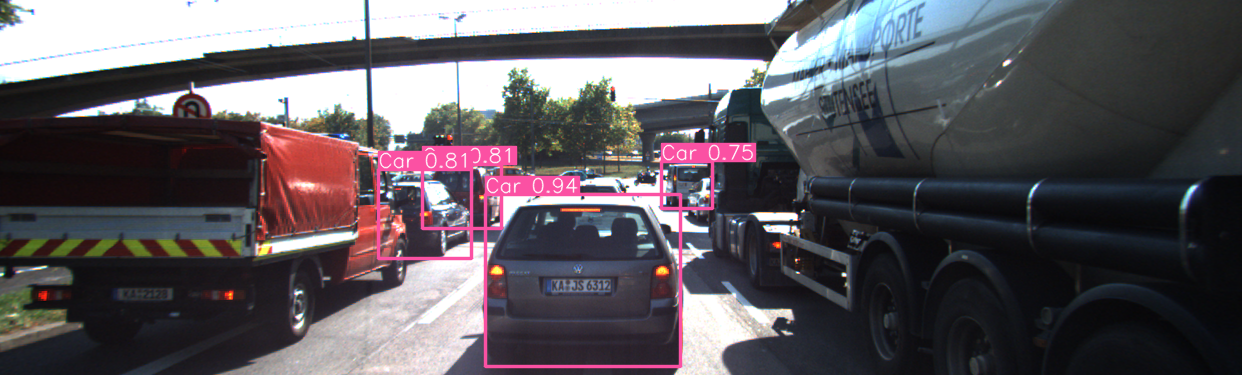}
        \centerline{(d) 1.YOLOV8n}
    \end{minipage}

    \begin{minipage}{\linewidth}
        \centering
        \includegraphics[width=\linewidth,height=4cm,keepaspectratio]{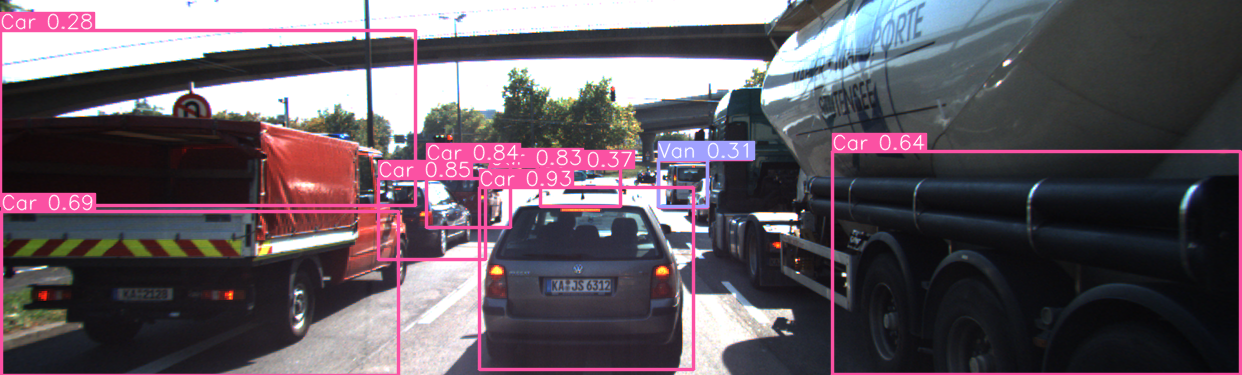}
        \centerline{(g) 1.YOLOV8s}
    \end{minipage}

    \begin{minipage}{\linewidth}
        \centering
        \includegraphics[width=\linewidth,height=4cm,keepaspectratio]{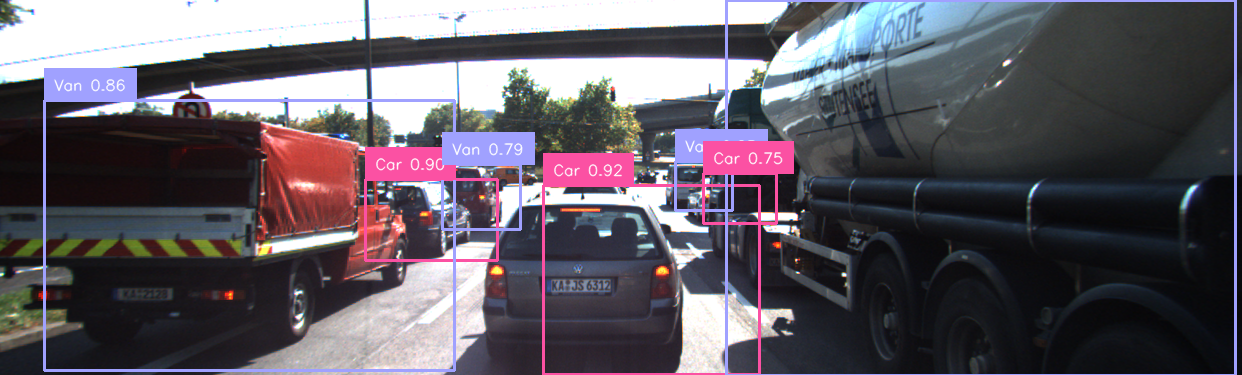}
        \centerline{(j) 1.YOLO-World}
    \end{minipage}

    \begin{minipage}{\linewidth}
        \centering
        \includegraphics[width=\linewidth,height=4cm,keepaspectratio]{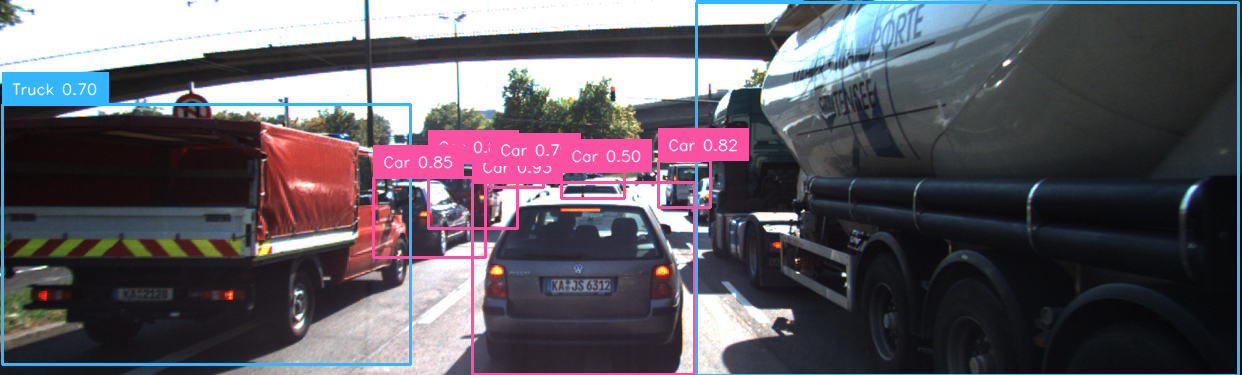}
        \centerline{(m) 1.YOLO-Vehicle-v1s}
    \end{minipage}
    
\end{minipage}%
\hfill
\begin{minipage}[t]{0.33\linewidth}
    \begin{minipage}{\linewidth}
        \centering
        \includegraphics[width=\linewidth,height=4cm,keepaspectratio]{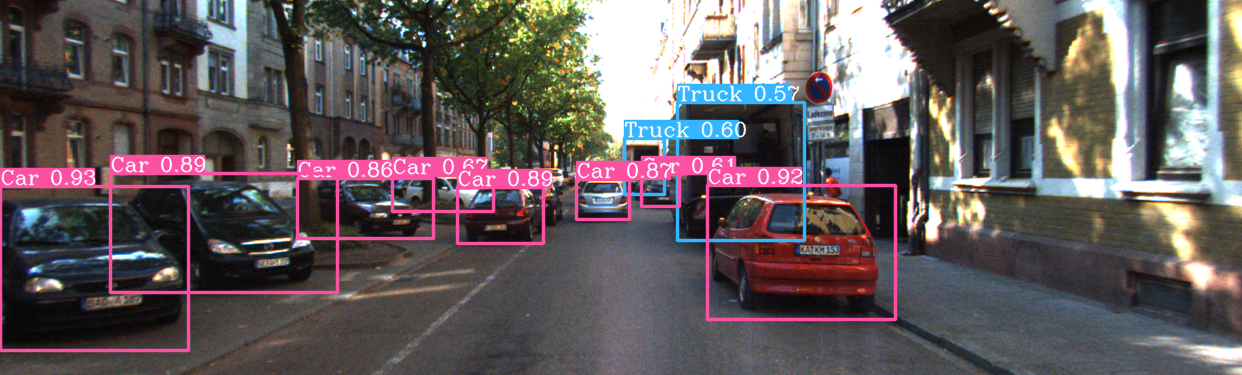}
        \centerline{(b) 2.YOLOV6-s6}
    \end{minipage}

    \begin{minipage}{\linewidth}
        \centering
        \includegraphics[width=\linewidth,height=4cm,keepaspectratio]{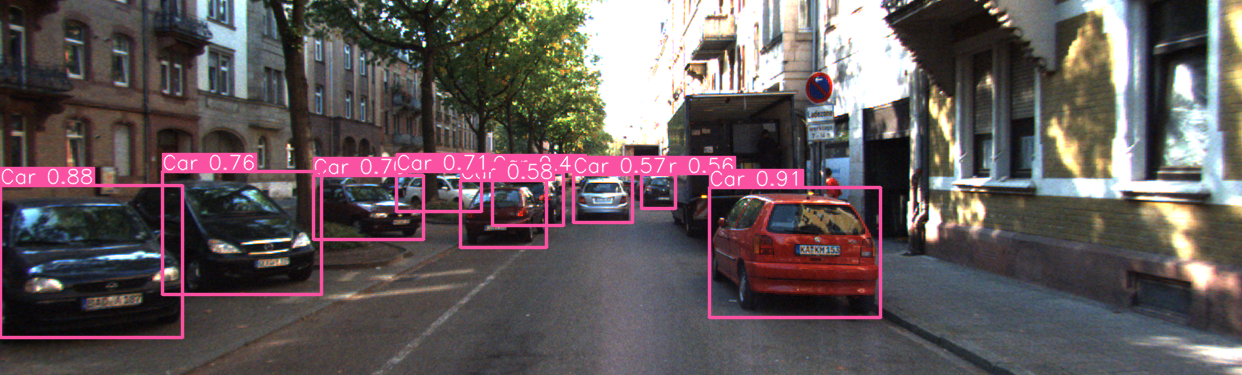}
        \centerline{(e) 2.YOLOV8n}
    \end{minipage}

    \begin{minipage}{\linewidth}
        \centering
        \includegraphics[width=\linewidth,height=4cm,keepaspectratio]{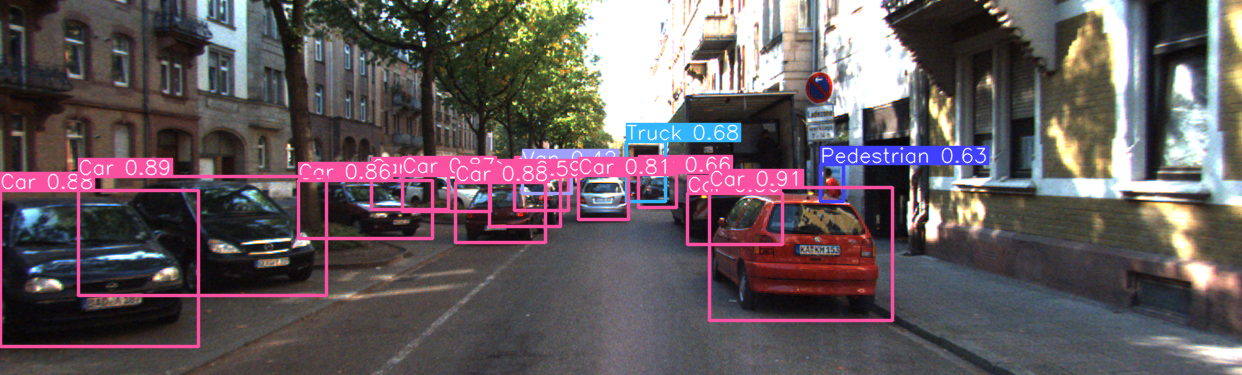}
        \centerline{(h) 2.YOLOV8s}
    \end{minipage}

    \begin{minipage}{\linewidth}
        \centering
        \includegraphics[width=\linewidth,height=4cm,keepaspectratio]{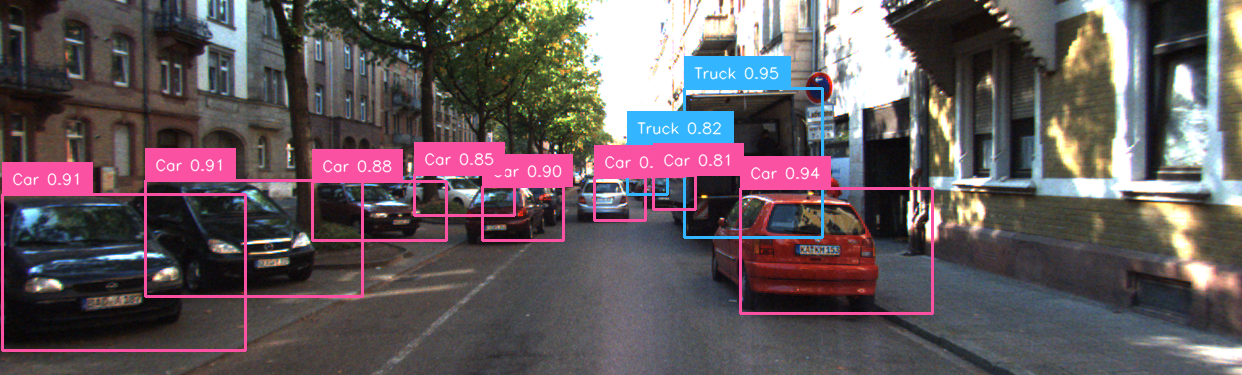}
        \centerline{(k) 2.YOLO-World}
    \end{minipage}
    
    \begin{minipage}{\linewidth}
        \centering
        \includegraphics[width=\linewidth,height=4cm,keepaspectratio]{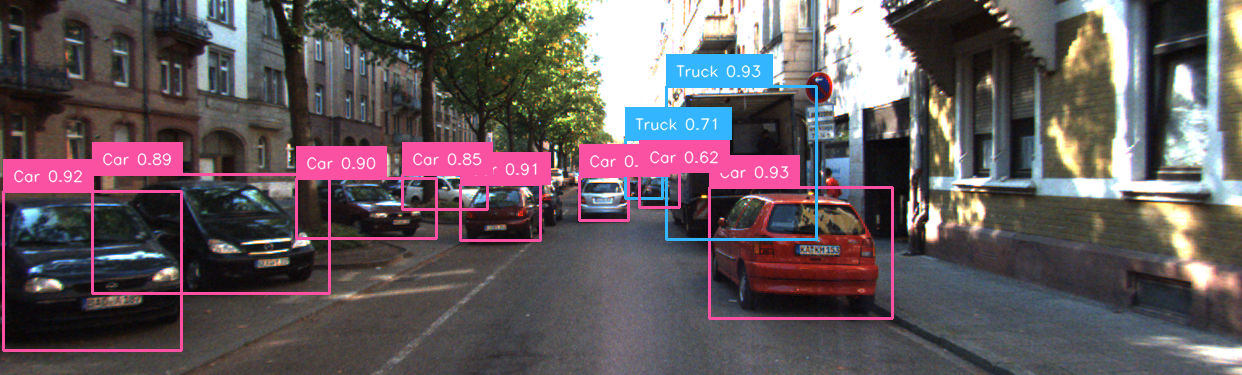}
        \centerline{(n) 2.YOLO-Vehicle-v1s}
    \end{minipage}
    
\end{minipage}%
\hfill
\begin{minipage}[t]{0.33\linewidth}
    \begin{minipage}{\linewidth}
        \centering
        \includegraphics[width=\linewidth,height=4cm,keepaspectratio]{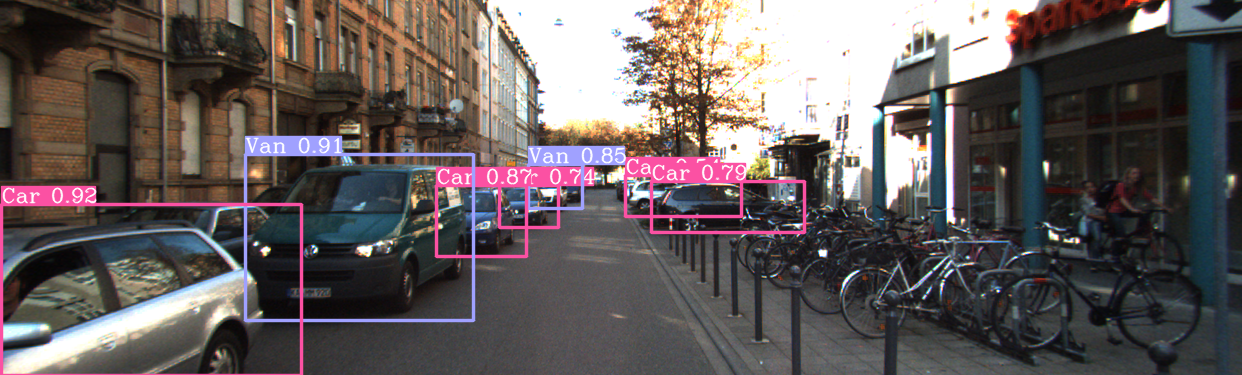}
        \centerline{(c) 3.YOLOV6-s6}
    \end{minipage}

    \begin{minipage}{\linewidth}
        \centering
        \includegraphics[width=\linewidth,height=4cm,keepaspectratio]{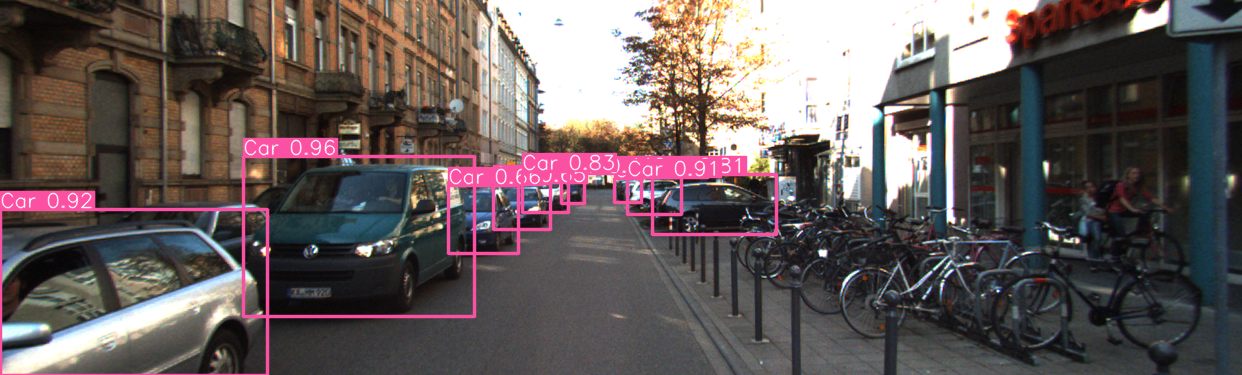}
        \centerline{(f) 3.YOLOV8n}
    \end{minipage}

    \begin{minipage}{\linewidth}
        \centering
        \includegraphics[width=\linewidth,height=4cm,keepaspectratio]{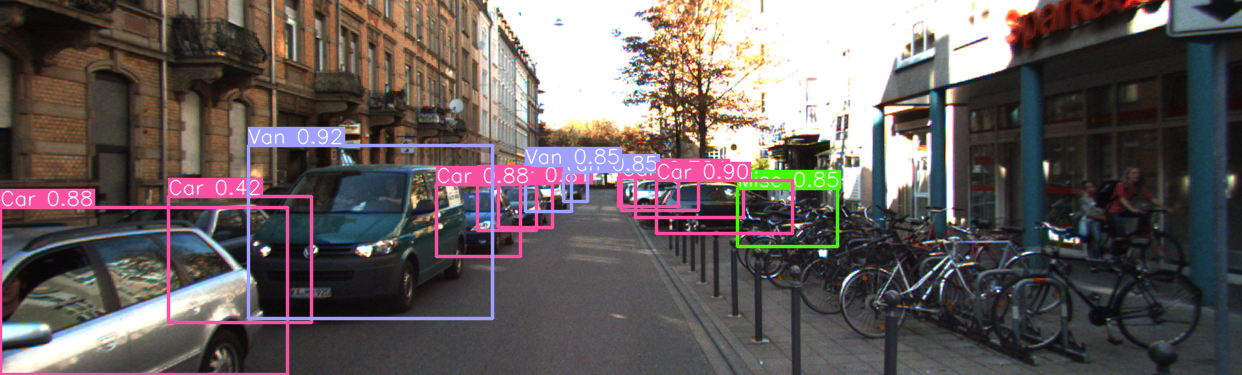}
        \centerline{(i) 3.YOLOV8s}
    \end{minipage}

    \begin{minipage}{\linewidth}
        \centering
        \includegraphics[width=\linewidth,height=4cm,keepaspectratio]{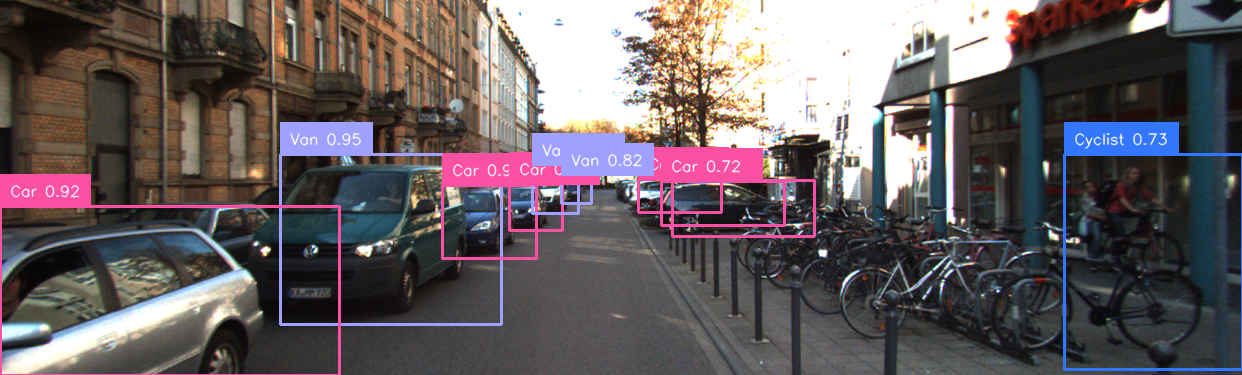}
        \centerline{(l) 3.YOLO-World}
    \end{minipage}

    \begin{minipage}{\linewidth}
        \centering
        \includegraphics[width=\linewidth,height=4cm,keepaspectratio]{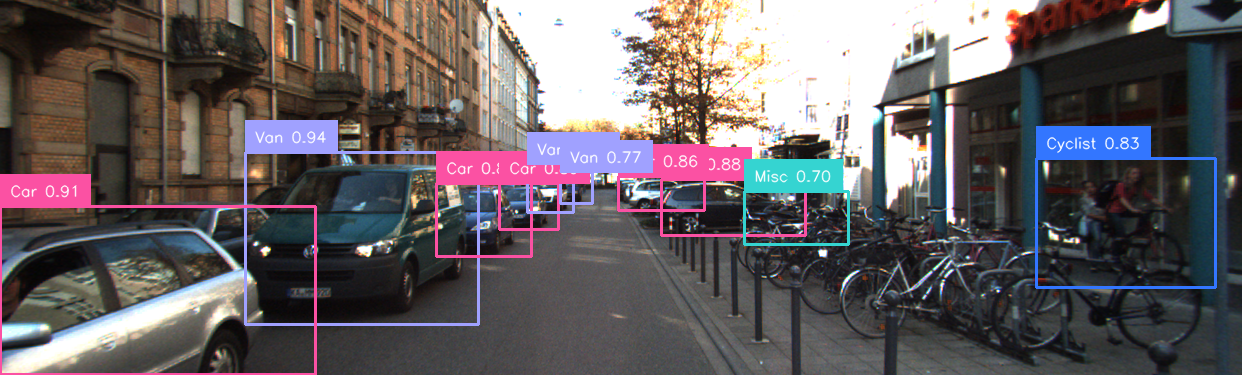}
        \centerline{(o) 3.YOLO-Vehicle-v1s}
    \end{minipage}
    
\end{minipage}
\end{minipage}
\caption{Performance of YOLO6-s6, YOLOV8n, YOLOV8s, YOLO-World, and YOLO-Vehicle models on the KITTI 2D detection test dataset. The YOLO-Vehicle model demonstrates excellent recognition capabilities, improving the detection ability for occluded objects and reducing the rates of false positives and false negatives.}
\label{fig:kitti_detection}
\end{figure*}

For the KITTI dataset, a similar training strategy to COCO2017 was adopted, with fine-tuning for dataset-specific characteristics. The model was trained on 7,481 training images, with a test dataset of 7,518 images. To fully leverage the KITTI dataset's features, the data augmentation strategy was adjusted by increasing the proportion of random cropping and rotation, simulating vehicle detection scenarios under various viewpoints and occlusions. Fig.~\ref{fig:kitti_loss} illustrates the LOSS function values of YOLOV6, YOLOV8n, YOLOV8s, YOLO-World, and YOLO-Vehicle models trained on the KITTI dataset. Fig.~\ref{fig:kitti_detection} presents the detection results of YOLOV6-s6, YOLOV8n, YOLOV8s, YOLO-World, and YOLO-Vehicle-v1s model on the KITTI dataset.

\begin{table*}[]
\centering
\caption{Comparison of the proposed method with previous object detection networks on the KITTI dataset}
\renewcommand{\arraystretch}{1.5}
\begin{tabular}{c|ccccccc}
\hline
\textbf{Model}            & \textbf{Input    Size} & \textbf{mAP}  & \textbf{mAP\textsubscript{@50}} & \textbf{mAP\textsubscript{@75}} & \textbf{FPS} & \textbf{Inference    Time} & \textbf{Size(MB)} \\ \hline
YOLOV6-s6        & 1280x1280     & 87.7      & 88.2   & 65.5   & 215    & 15ms              & 90       \\\hline
YOLOV8n          & 1280x1280     & 85.6      & 89.4   & 72.0   & 425.61 & 7.9ms             & 10       \\\hline
YOLOV8s          & 1280x1280     & 89.2      & 89.3   & 72.1   & 264.42 & 9ms               & 23       \\\hline
Edge YOLO{[}1{]} &               &           & 82.1   &        &        &                   &          \\\hline
YOLO-World       & 1280x1280     & 89.0      & 90.0   & 77.1   & 93     & 34ms              & 700      \\\hline
\textbf{YOLO-Vehicle-v1s} & 1280x384      & \textbf{92.1}      & \textbf{93.5}   & \textbf{81.1}   & \textbf{226}    & \textbf{12ms}              & \textbf{45.4}    \\\hline
\end{tabular}
\label{tab:kitti_table}
\end{table*}

The results in Table~\ref{tab:kitti_table} demonstrate that the YOLO-Vehicle-v1s model performs excellently across multiple key metrics. In terms of detection accuracy, the model achieves mAP of 92.1\%, with mAP\textsubscript{@50} and mAP\textsubscript{@75} at 93.5\% and 81.1\% respectively, outperforming other models in the comparison table. This indicates that YOLO-Vehicle-v1s maintains high detection accuracy at different IoU thresholds. Regarding processing speed, the model reaches 226 FPS, second only to YOLOV8n, but significantly superior to YOLO-World and YOLOV6 models. Notably, YOLO-Vehicle-v1s achieves model lightweight while maintaining high performance. Its model size is 45.4MB, 15 times smaller than YOLO-World's 700MB model size. To verify the model's adaptability on different computing platforms, this paper also conducted deployment tests on embedded devices (such as Jetson Nano). Results show that even in resource-constrained environments, YOLO-Vehicle-v1s can maintain a processing speed close to 30 FPS with only a slight decrease in detection accuracy.

Furthermore, to address the challenge of recognition in hazy conditions, this paper proposes an improved model based on YOLO-Vehicle, called YOLO-Vehicle-Pro, which can efficiently detect objects in hazy weather. The Foggy Cityscapes dataset introduced in Table~\ref{tab:datasets} was selected, using 1000 images from the training dataset and 500 images from the test dataset, which were input into the Ucl-Dehaze model \cite{19} for training. The parameter settings are as follows: n\_epochs is set to 50. In the first 50 epochs, training is conducted using the initial learning rate. n\_epochs\_decay is set to 500, and during epochs 50-500, the learning rate gradually decreases to zero for training. The initial learning rate lr is set to 0.0002. gan\_mode is set to vanilla, using the cross-entropy objective. lr\_policy is set to cosine, implementing cosine annealing.

\begin{figure}[htbp]
    \centering
    \includegraphics[width=0.5\textwidth]{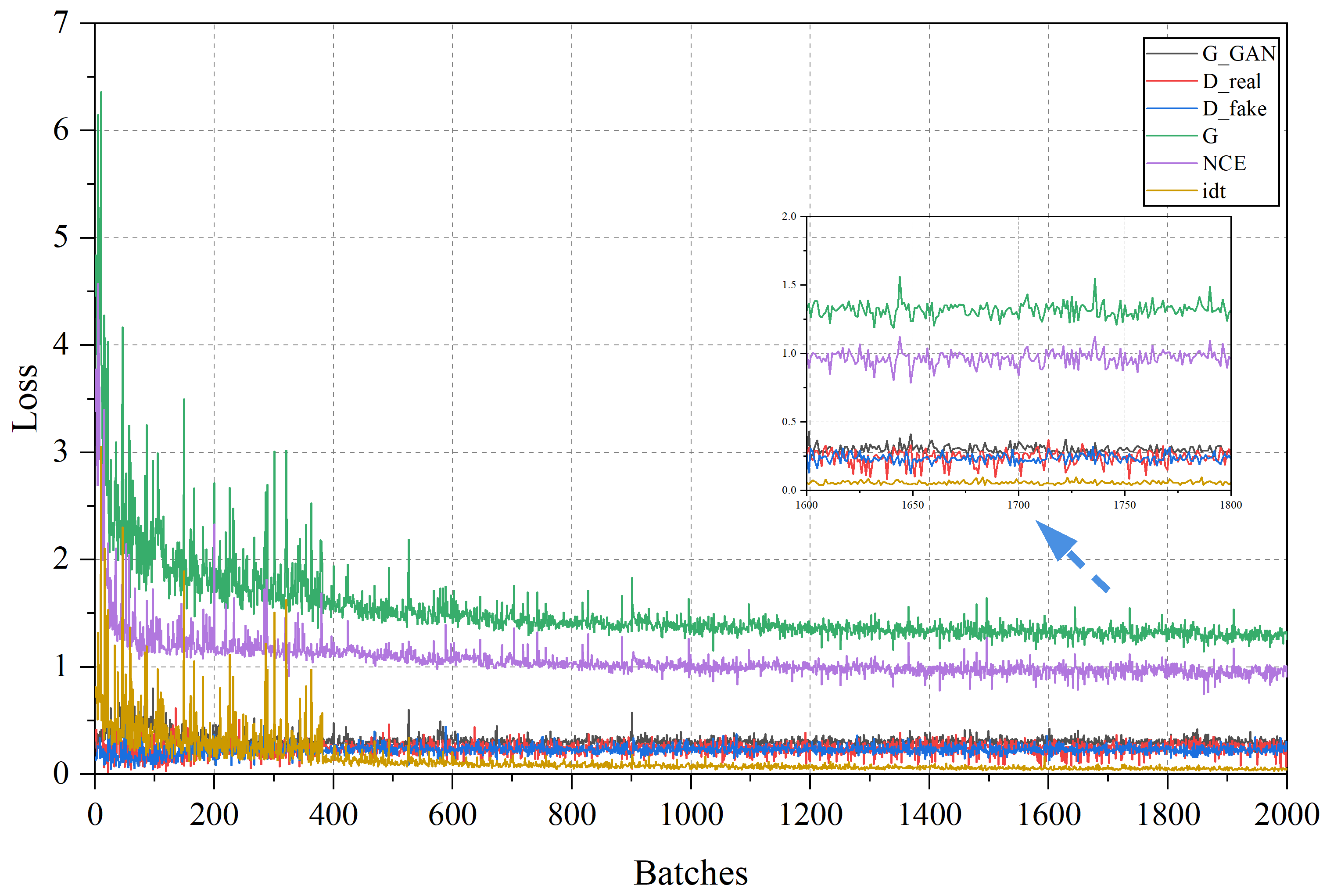}
    \caption{Trends of various loss functions during the training process of the dehazing module}
    \label{fig:loss3}
\end{figure}

Fig.~\ref{fig:loss3} illustrates the variations of different loss functions with respect to training epochs during the dehazing module training process. The graph contains multiple curves, specifically: G\_GAN represents the adversarial loss of the generator, D\_real is the discriminator's loss for real images, D\_fake is the discriminator's loss for generated images, G is the overall loss of the generator, NCE is the contrastive learning loss, idt is the identity loss, and perceptual is the perceptual loss. The horizontal axis represents the number of training epochs, while the vertical axis indicates the loss values. From the graph, it can be observed that as training progresses, all loss components show a general downward trend, indicating continuous model optimization during the training process. Among these, the generator loss(G) and contrastive learning loss(NCE) exhibit larger fluctuations, but their overall trends are decreasing, reflecting the dynamic adjustment process of the generator and contrastive learning components during training.

\begin{figure}[htbp]
\centering 
\begin{minipage}[t]{0.5\textwidth}
\begin{minipage}[t]{0.5\linewidth}
    \begin{minipage}{\linewidth}
        \centering
        \includegraphics[width=\linewidth,height=3cm,keepaspectratio]{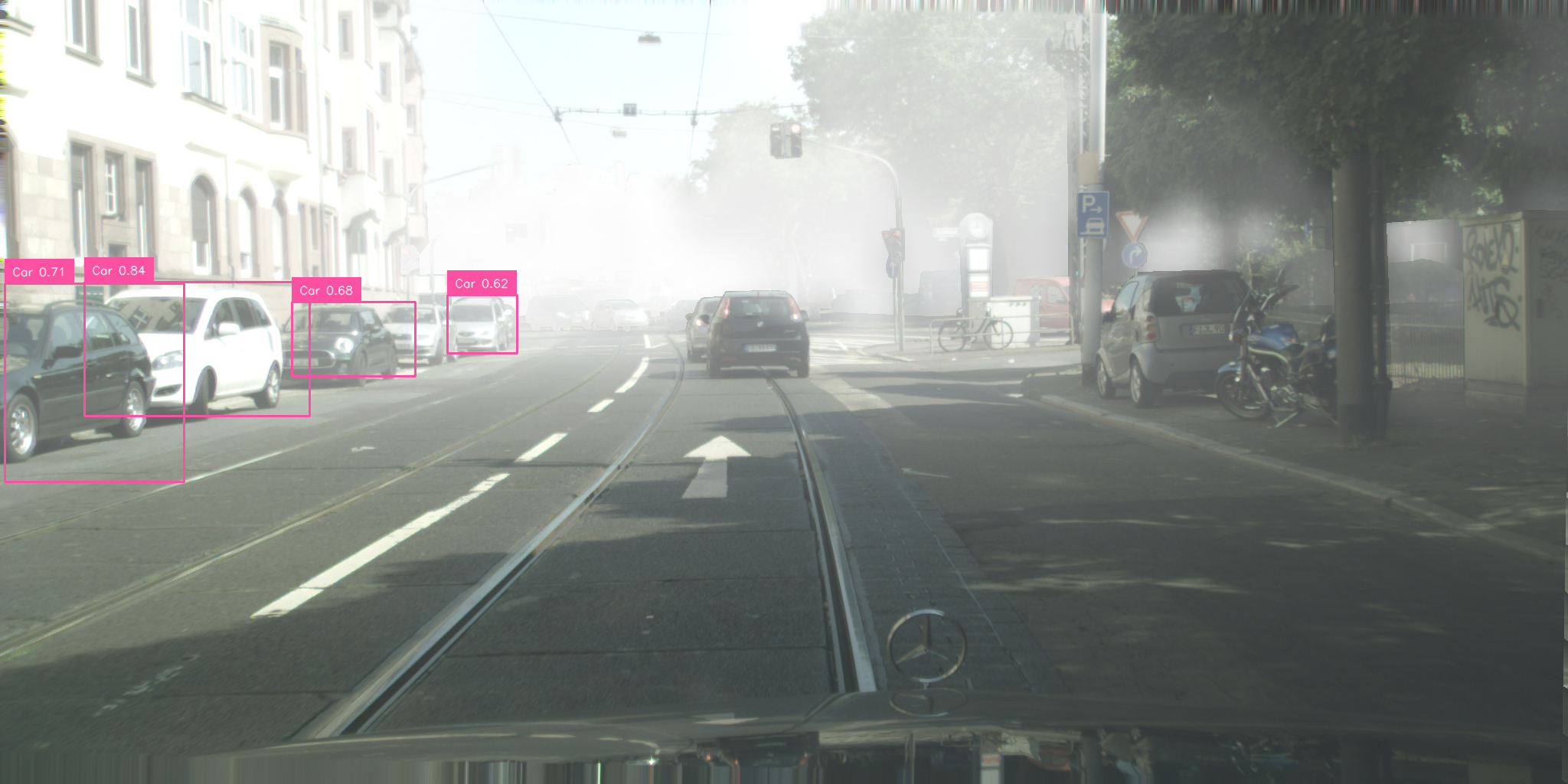}
        \centerline{(a) Foggy Cityscapes Test 1}
        \label{fig:val_007}
    \end{minipage}
    \begin{minipage}{\linewidth}
        \centering
        \includegraphics[width=\linewidth,height=3cm,keepaspectratio]{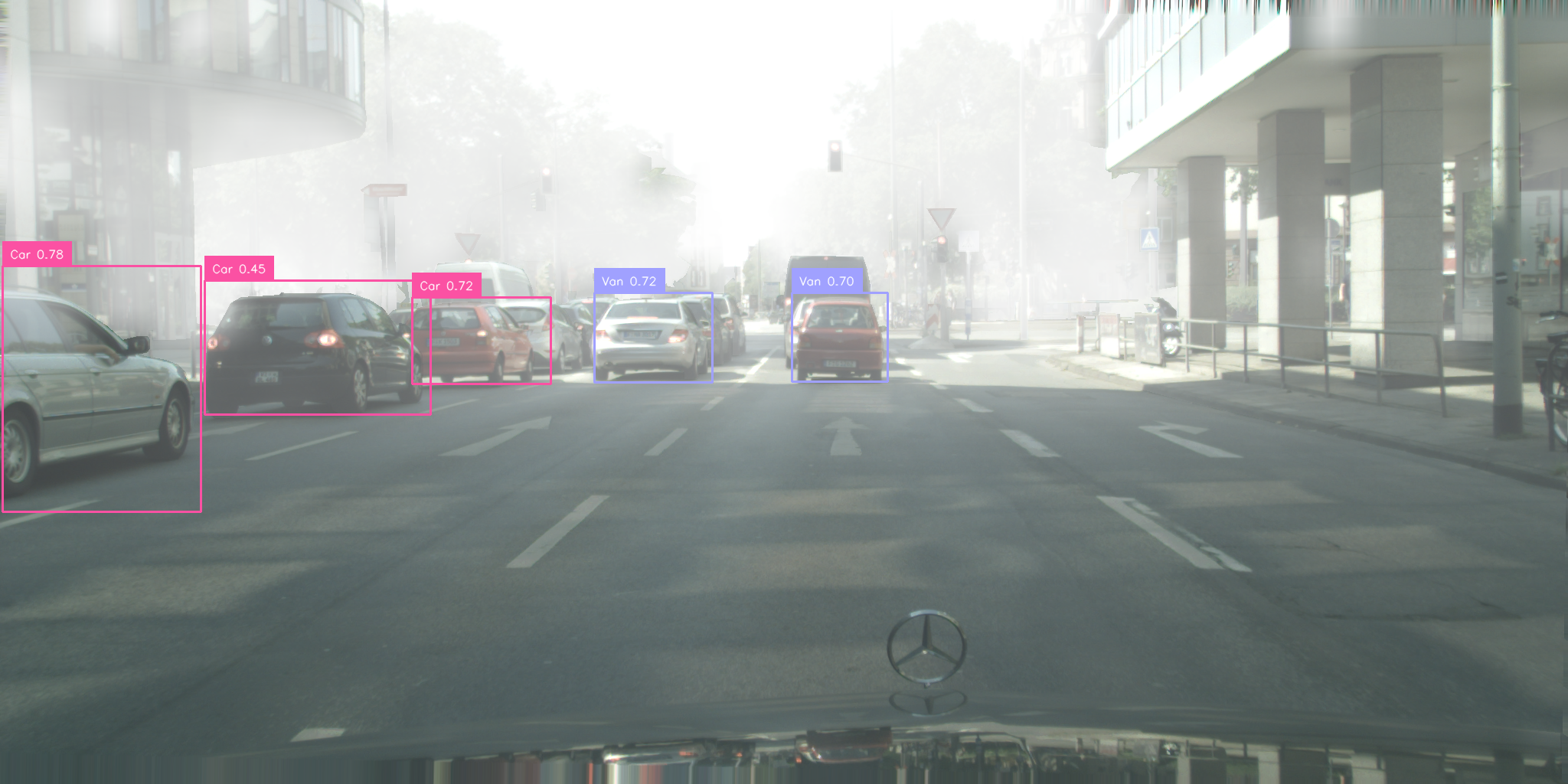}
        \centerline{(b) Foggy Cityscapes Test 2}
        \label{fig:val_115}
    \end{minipage}
\end{minipage}%
\hfill
\begin{minipage}[t]{0.5\linewidth}
    \begin{minipage}{\linewidth}
        \centering
        \includegraphics[width=\linewidth,height=3cm,keepaspectratio]{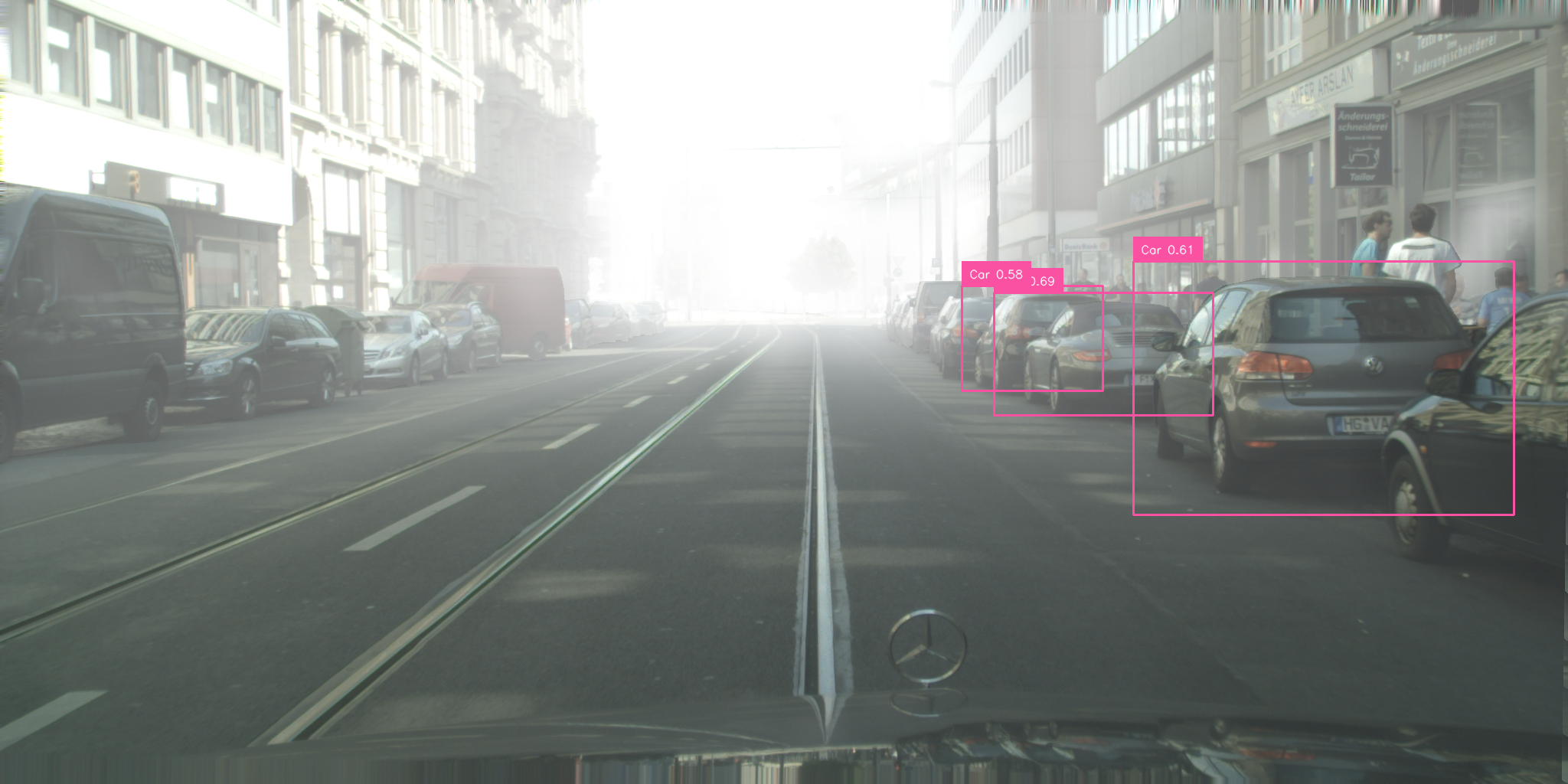}
        \centerline{(c) Foggy Cityscapes Test 3}
        \label{fig:val_136}
    \end{minipage}
    \begin{minipage}{\linewidth}
        \centering
        \includegraphics[width=\linewidth,height=3cm,keepaspectratio]{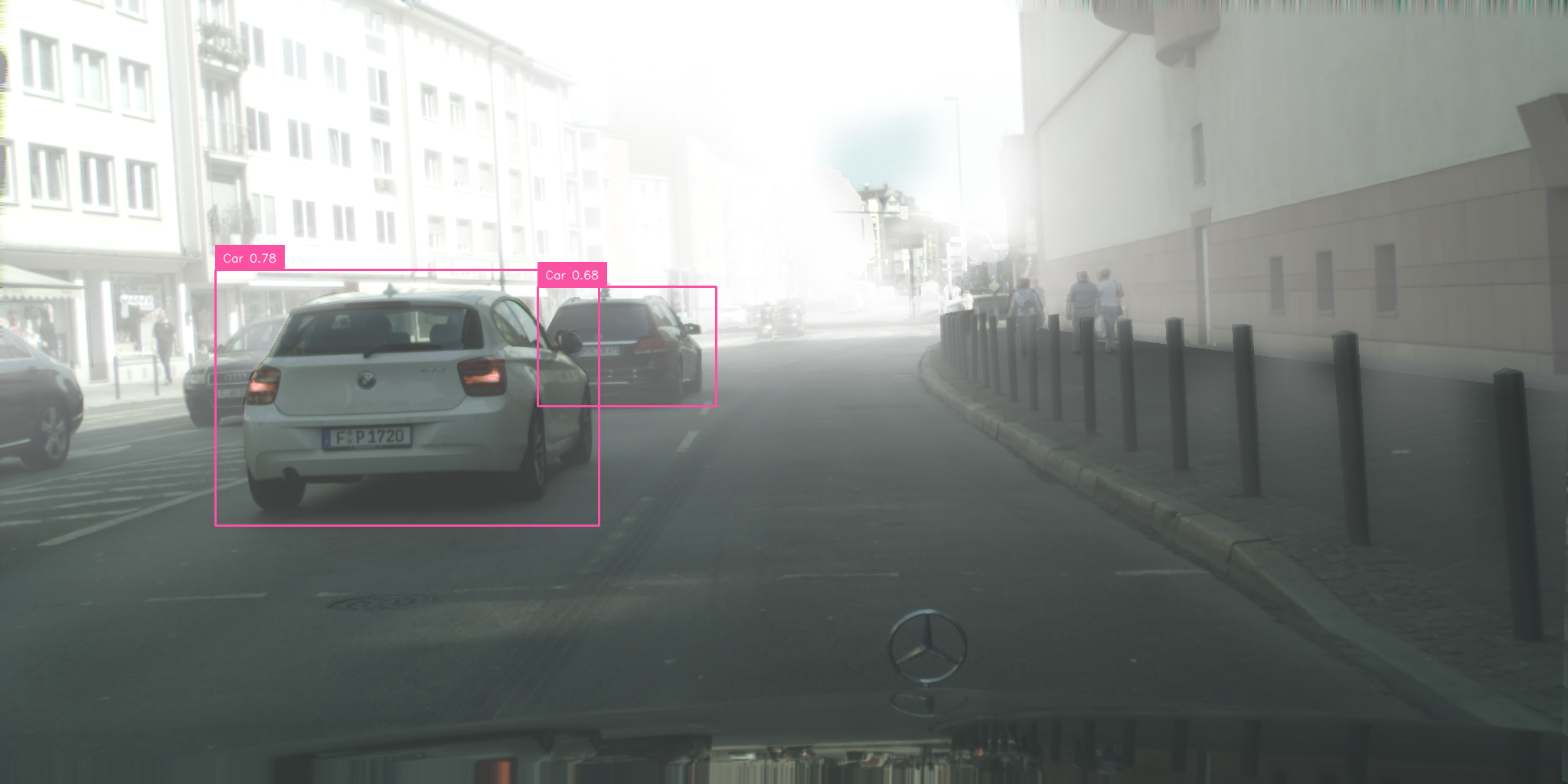}
        \centerline{(d) Foggy Cityscapes Test 4}
        \label{fig:val_206}
    \end{minipage}
\end{minipage}
\end{minipage}
\caption{Performance of YOLO-World on the Foggy Cityscapes detection test dataset}
\label{fig:yolo_world_haze}
\end{figure}

\begin{figure*}[htbp]
\centering 
\begin{minipage}[t]{0.98\textwidth}
\begin{minipage}[t]{0.33\linewidth}
    \begin{minipage}{0.98\linewidth}
        \centering
        \includegraphics[width=\linewidth,height=2.6cm,keepaspectratio]{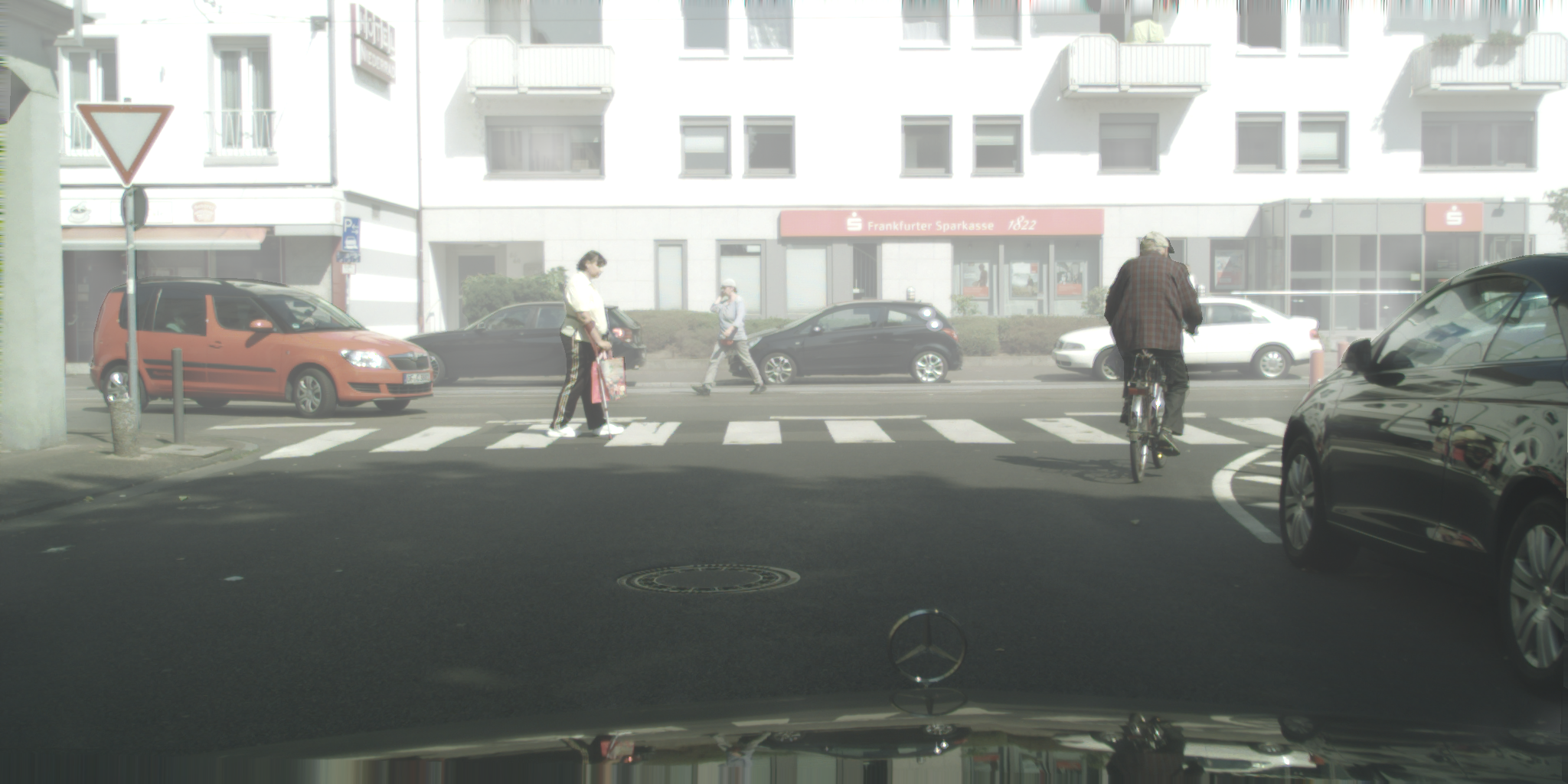}
    \end{minipage}

    \vspace{-0.02cm}
    \begin{minipage}{0.98\linewidth}
        \centering
        \includegraphics[width=\linewidth,height=2.6cm,keepaspectratio]{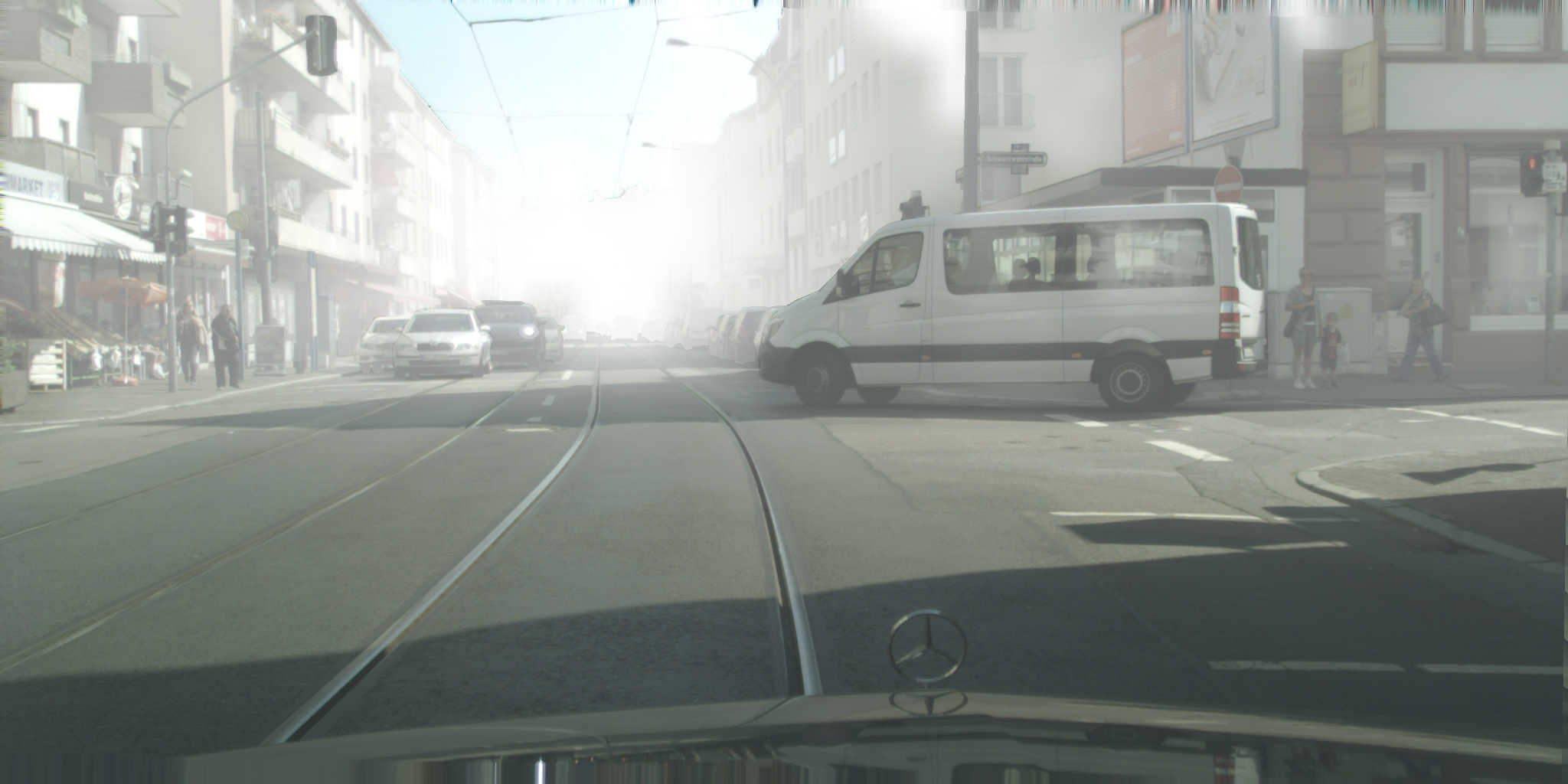}
    \end{minipage}

    \vspace{-0.2cm}
    \begin{minipage}{0.98\linewidth}
        \centering
        \includegraphics[width=\linewidth,height=2.6cm,keepaspectratio]{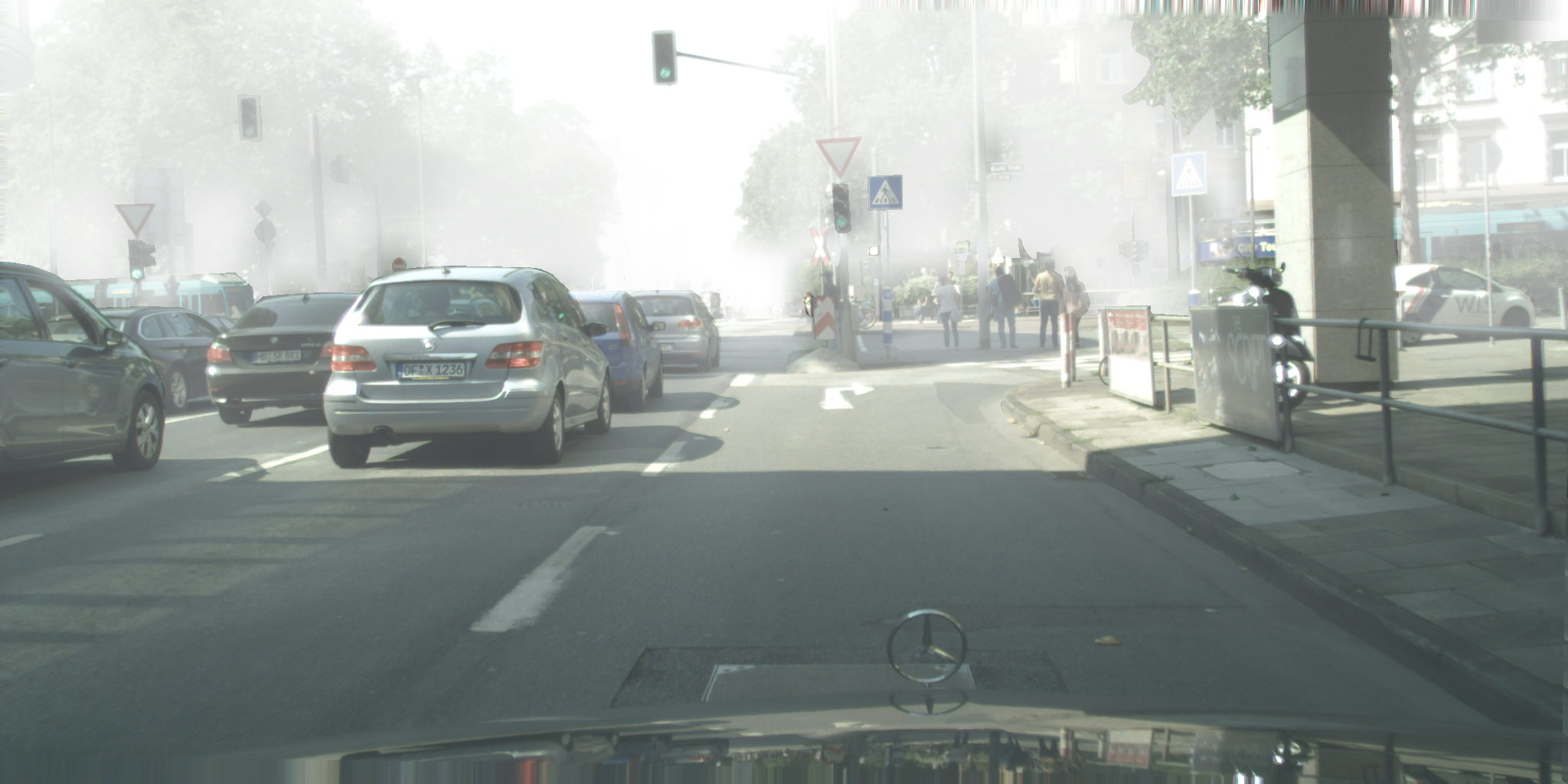}
    \end{minipage}

    \vspace{-0.2cm}
    \begin{minipage}{0.98\linewidth}
        \centering
        \includegraphics[width=\linewidth,height=2.6cm,keepaspectratio]{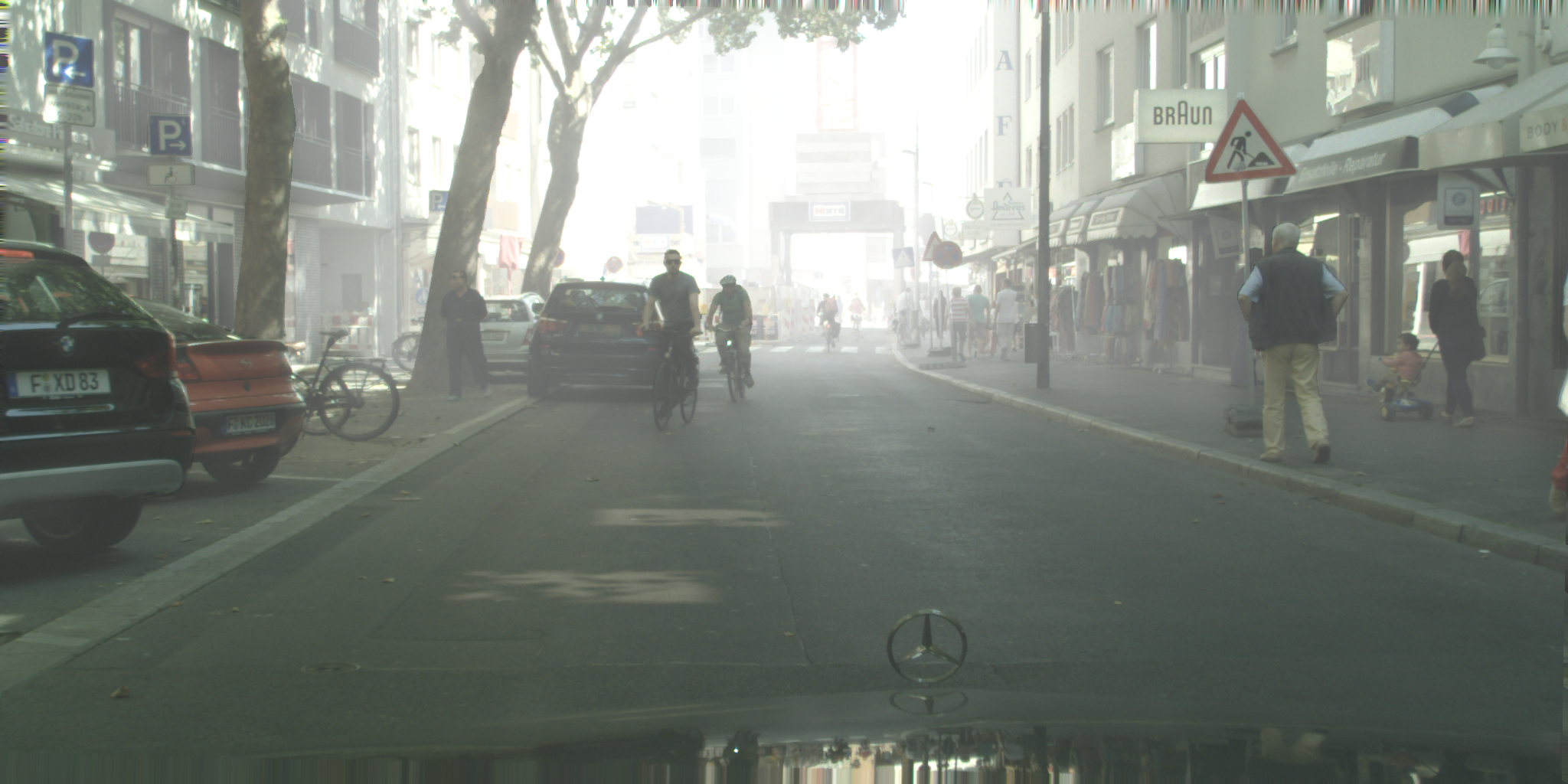}
    \end{minipage}

    \vspace{-0.2cm}
    \begin{minipage}{0.98\linewidth}
        \centering
        \includegraphics[width=\linewidth,height=2.6cm,keepaspectratio]{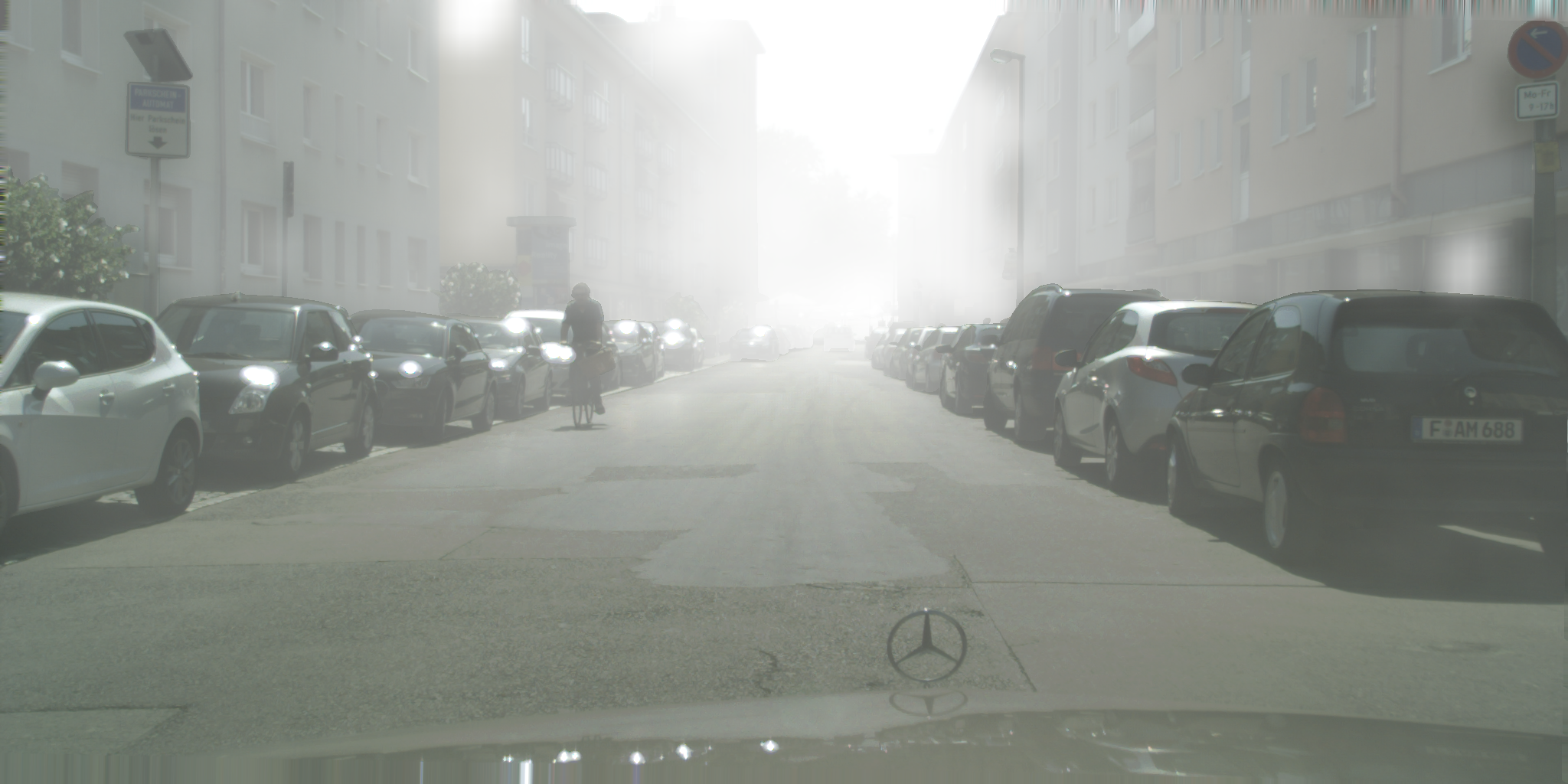}
        \centerline{(a) Real-World Hazy Images}
    \end{minipage}
\end{minipage}%
\hspace{-0.2cm}
\begin{minipage}[t]{0.33\linewidth}
    \begin{minipage}{0.98\linewidth}
        \centering
        \includegraphics[width=\linewidth,height=2.6cm,keepaspectratio]{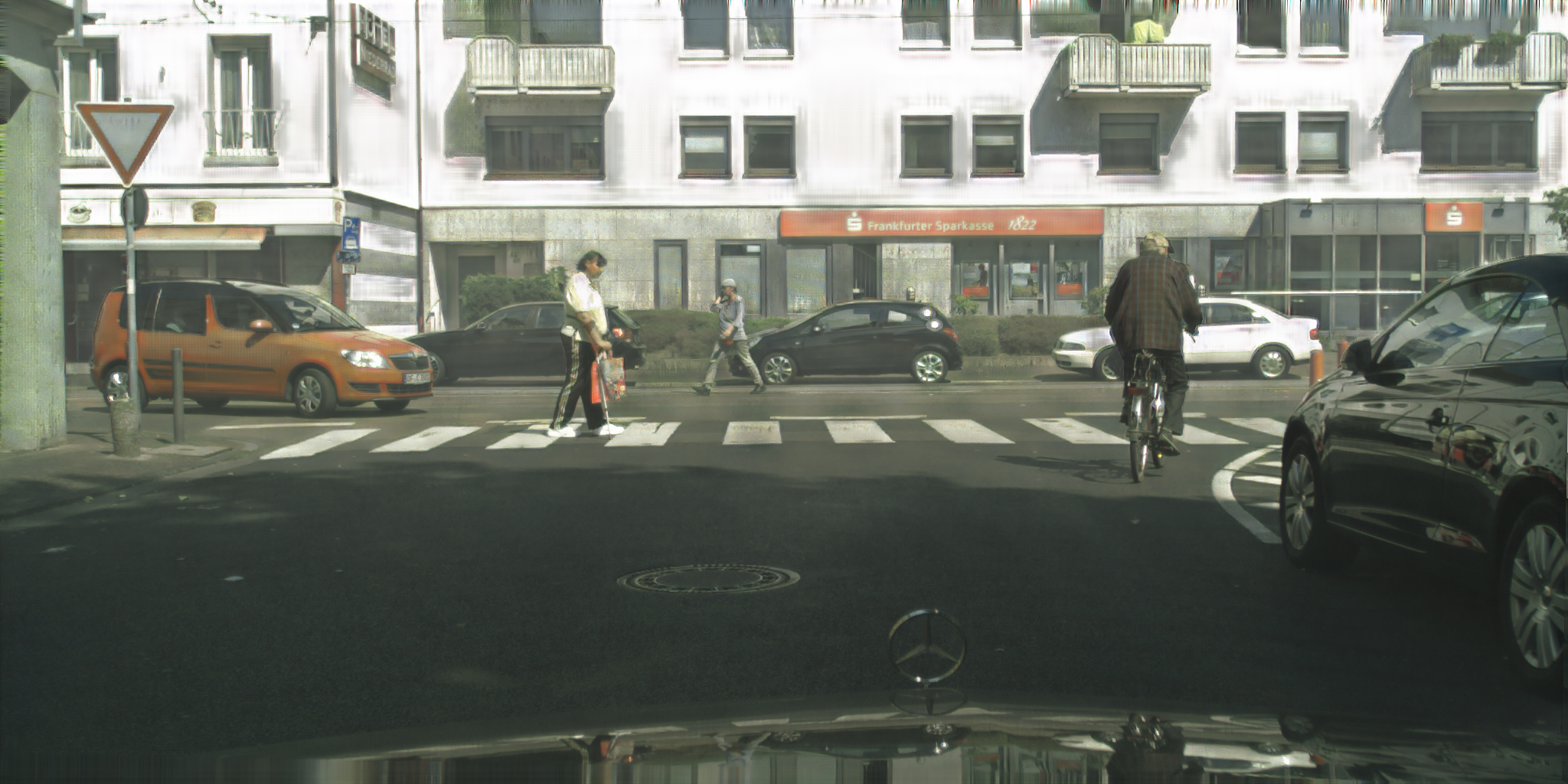}
    \end{minipage}

    \vspace{-0.02cm}
    \begin{minipage}{0.98\linewidth}
        \centering
        \includegraphics[width=\linewidth,height=2.6cm,keepaspectratio]{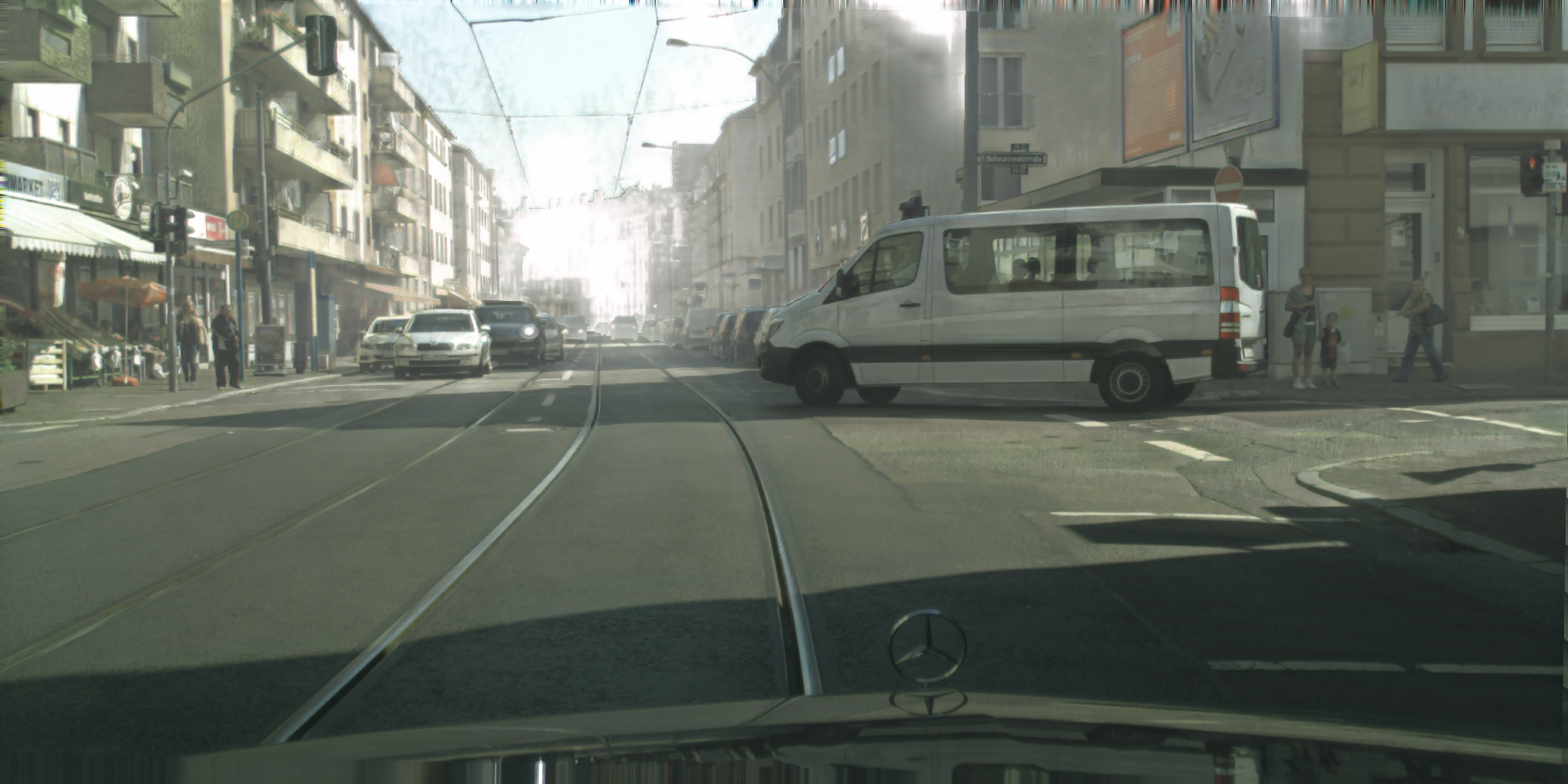}
    \end{minipage}

    \vspace{-0.2cm}
    \begin{minipage}{0.98\linewidth}
        \centering
        \includegraphics[width=\linewidth,height=2.6cm,keepaspectratio]{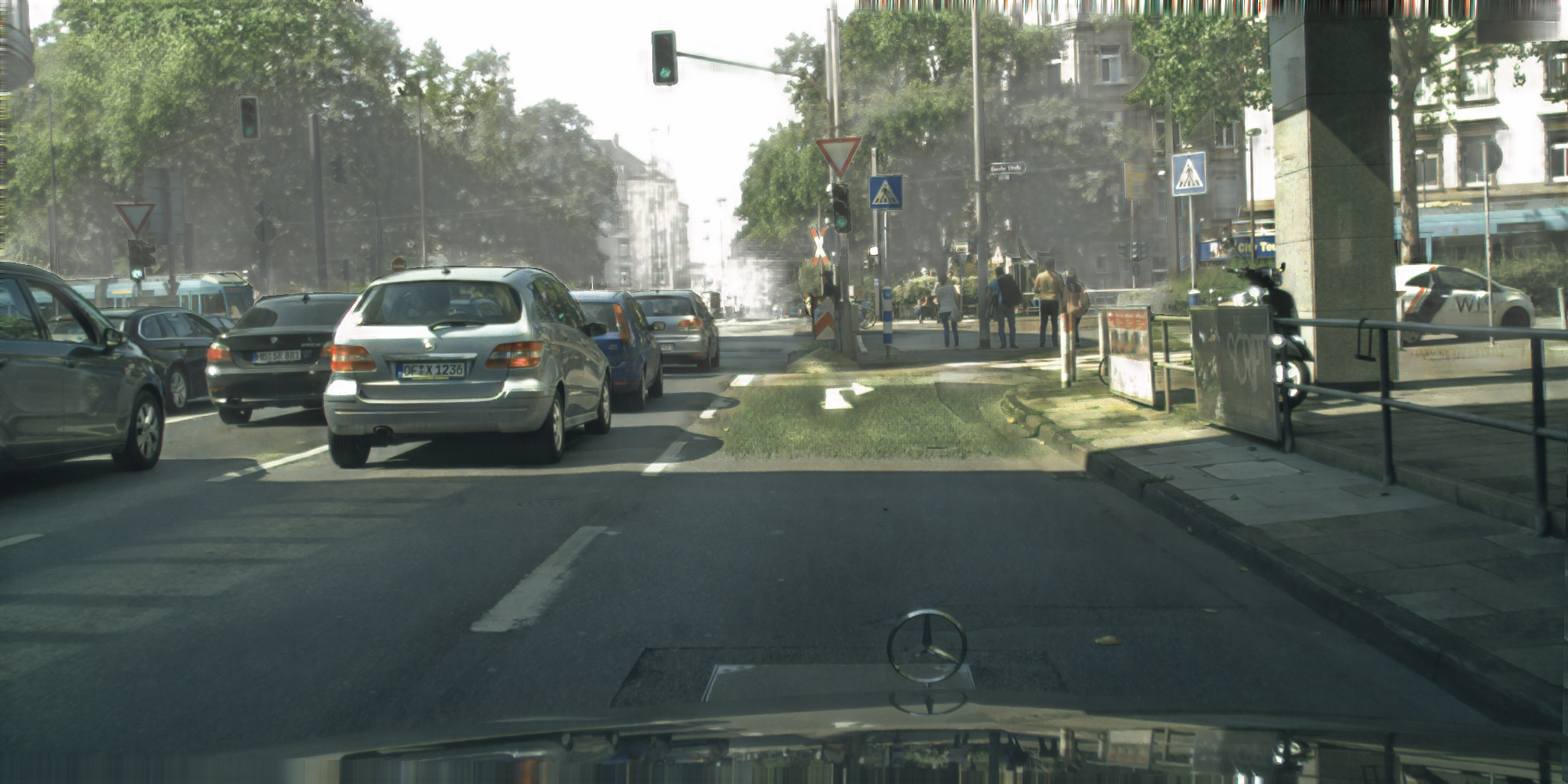}
    \end{minipage}

    \vspace{-0.2cm}
    \begin{minipage}{0.98\linewidth}
        \centering
        \includegraphics[width=\linewidth,height=2.6cm,keepaspectratio]{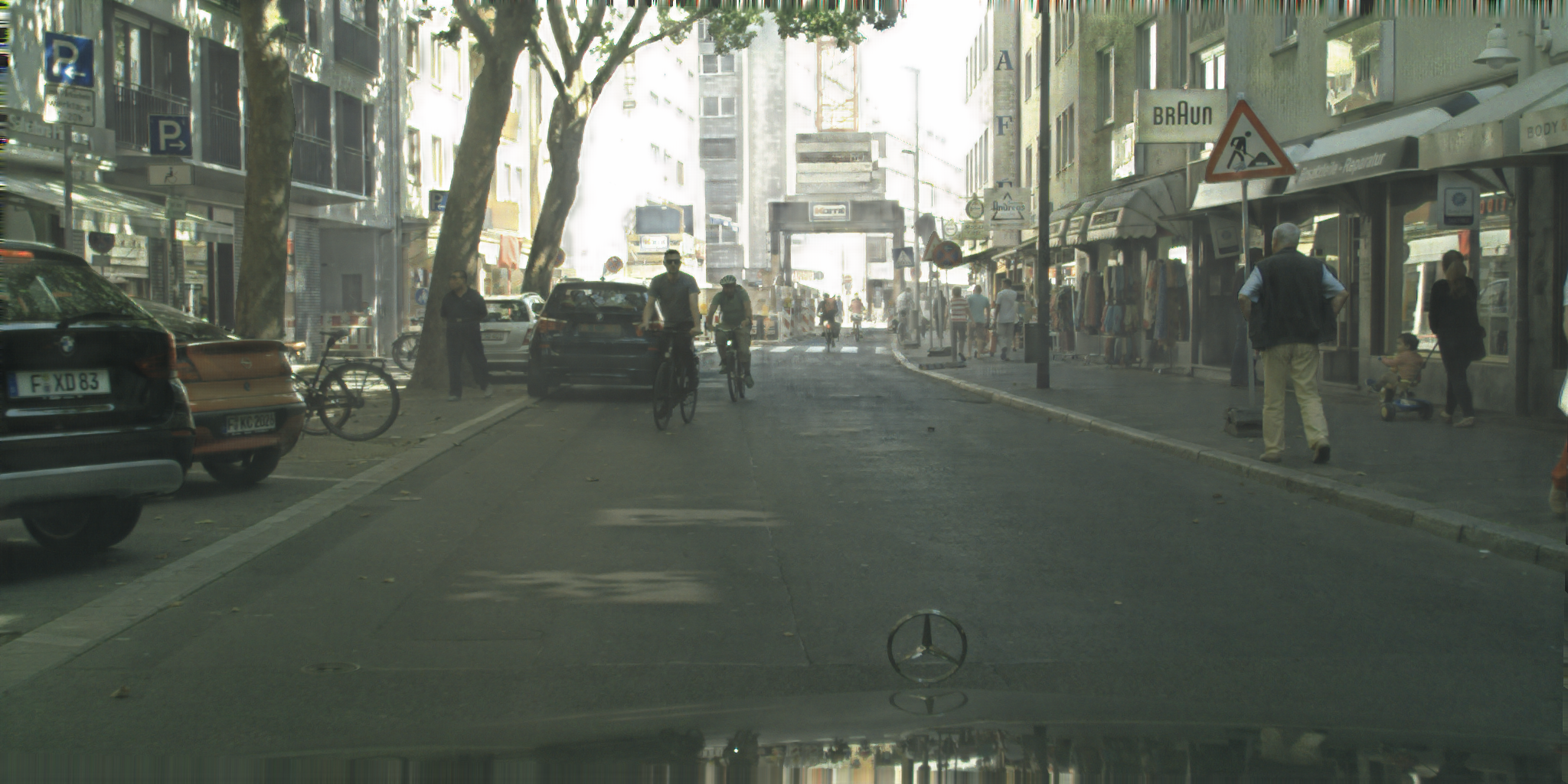}
    \end{minipage}

    \vspace{-0.2cm}
    \begin{minipage}{0.98\linewidth}
        \centering
        \includegraphics[width=\linewidth,height=2.6cm,keepaspectratio]{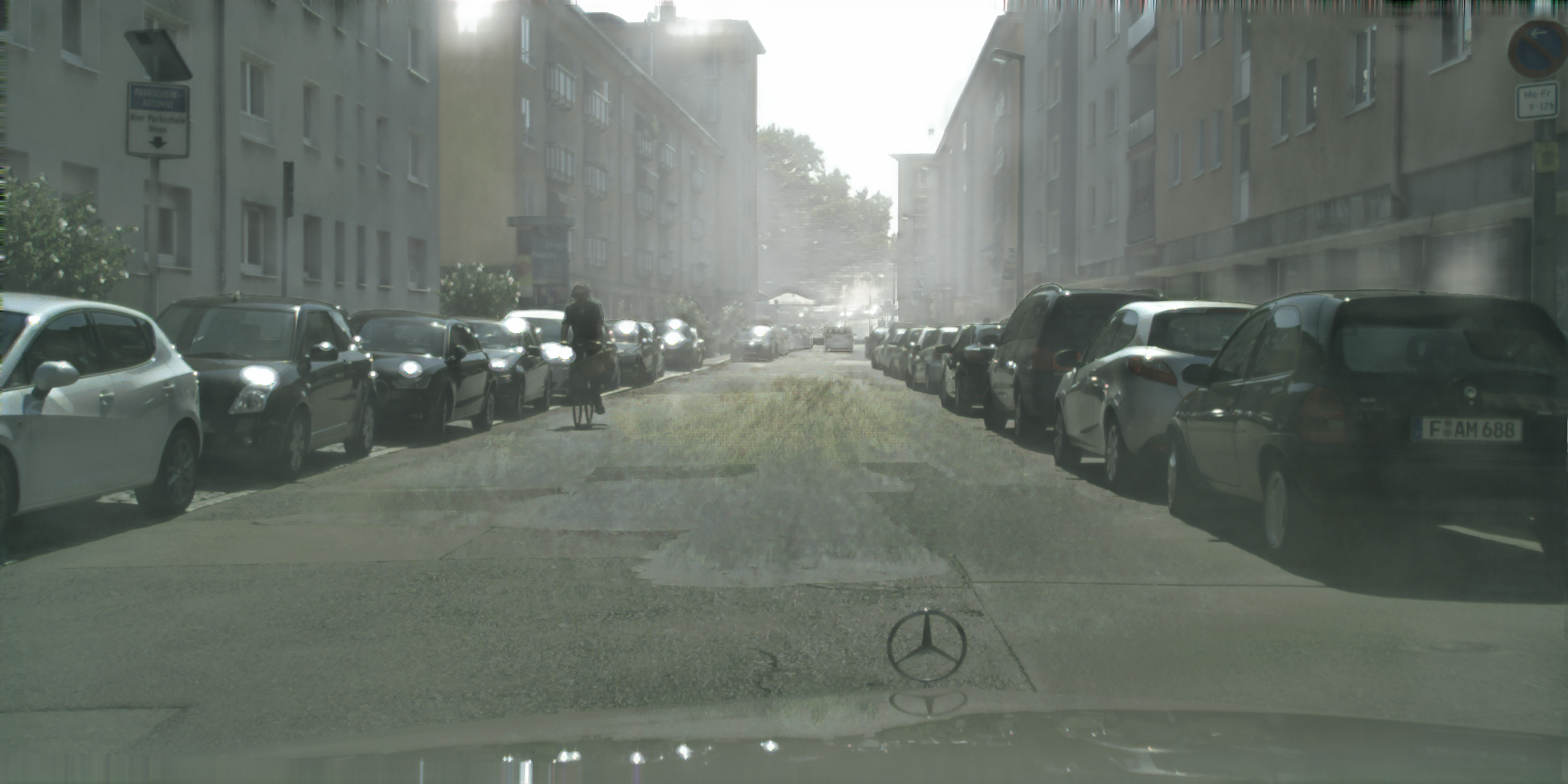}
        \centerline{(b) Dehazy Images}
    \end{minipage}
\end{minipage}%
\hspace{-0.2cm}
\begin{minipage}[t]{0.33\linewidth}
    \begin{minipage}{0.98\linewidth}
        \centering
        \includegraphics[width=\linewidth,height=2.6cm,keepaspectratio]{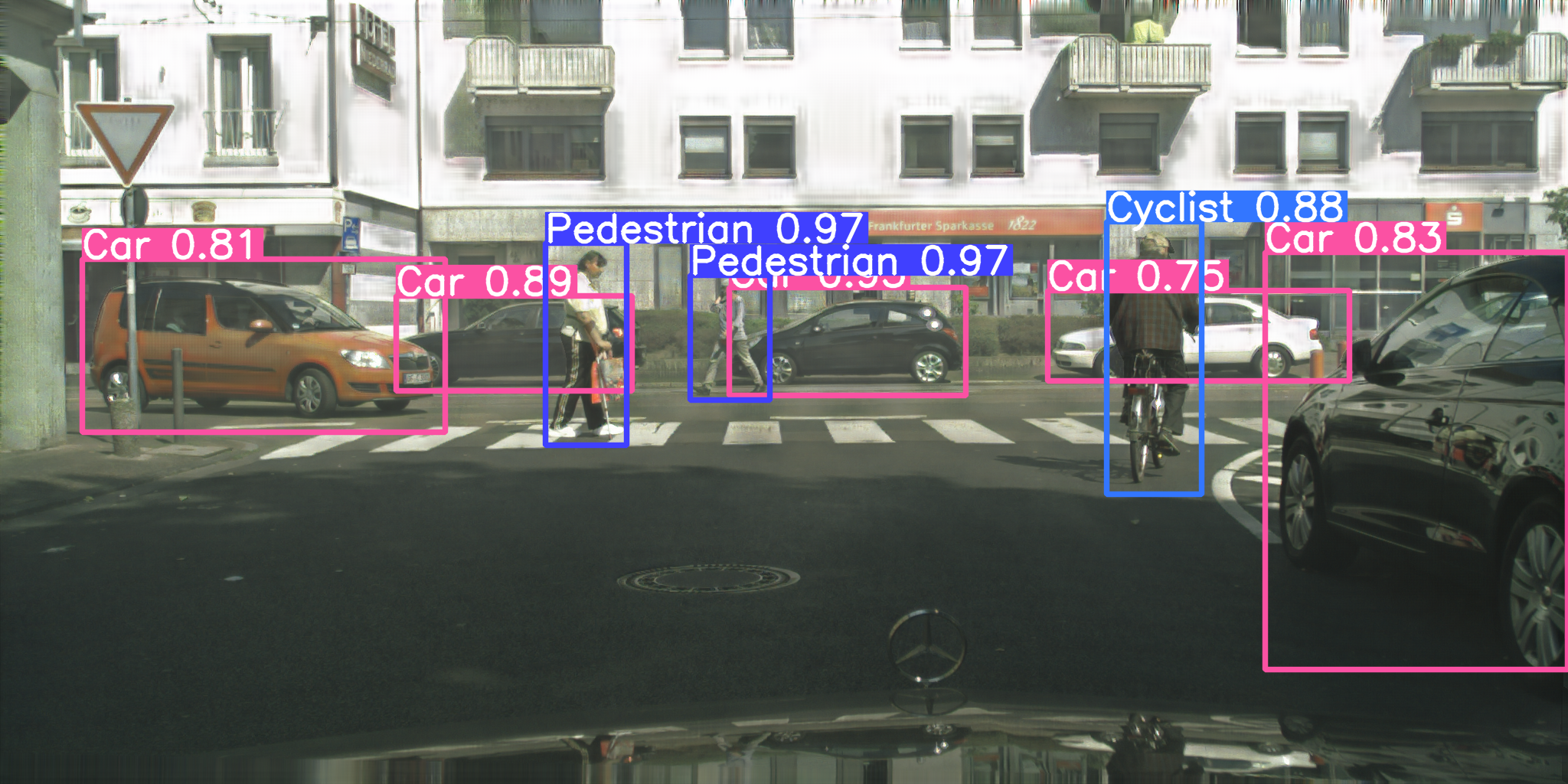}
    \end{minipage}

    \vspace{-0.02cm}
    \begin{minipage}{0.98\linewidth}
        \centering
        \includegraphics[width=\linewidth,height=2.6cm,keepaspectratio]{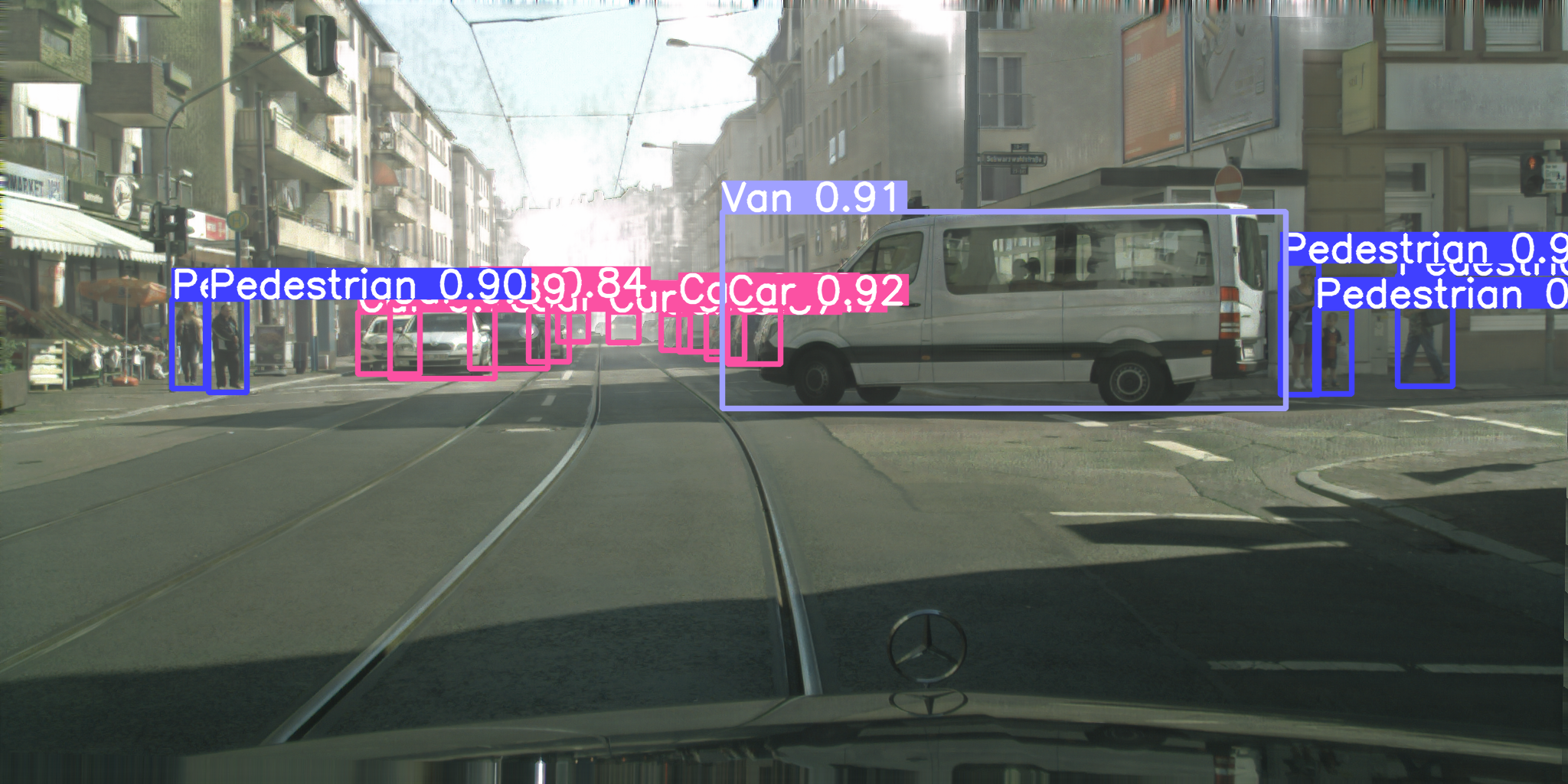}
    \end{minipage}

    \vspace{-0.2cm}
    \begin{minipage}{0.98\linewidth}
        \centering
        \includegraphics[width=\linewidth,height=2.6cm,keepaspectratio]{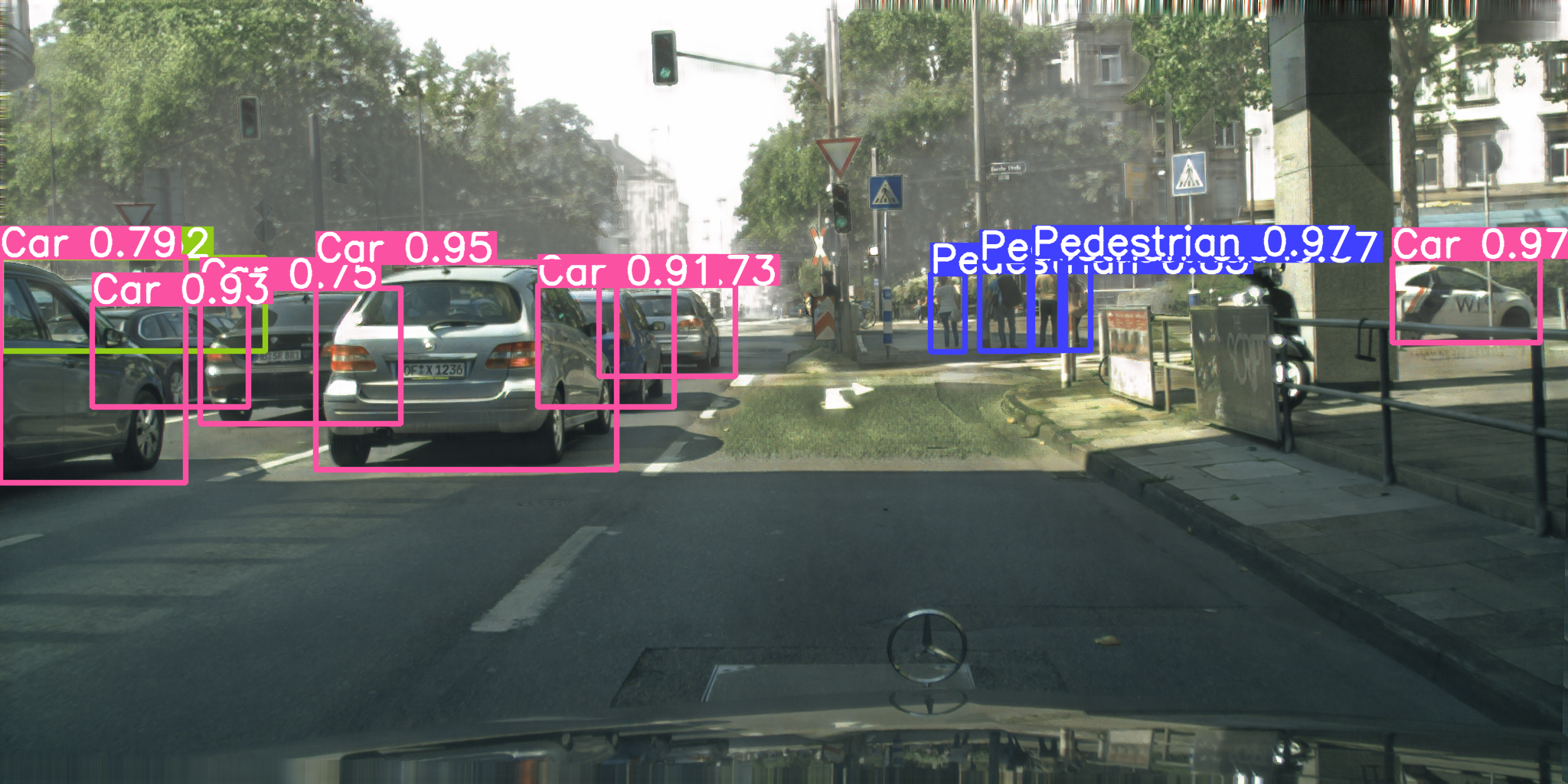}
    \end{minipage}

    \vspace{-0.2cm}
    \begin{minipage}{0.98\linewidth}
        \centering
        \includegraphics[width=\linewidth,height=2.6cm,keepaspectratio]{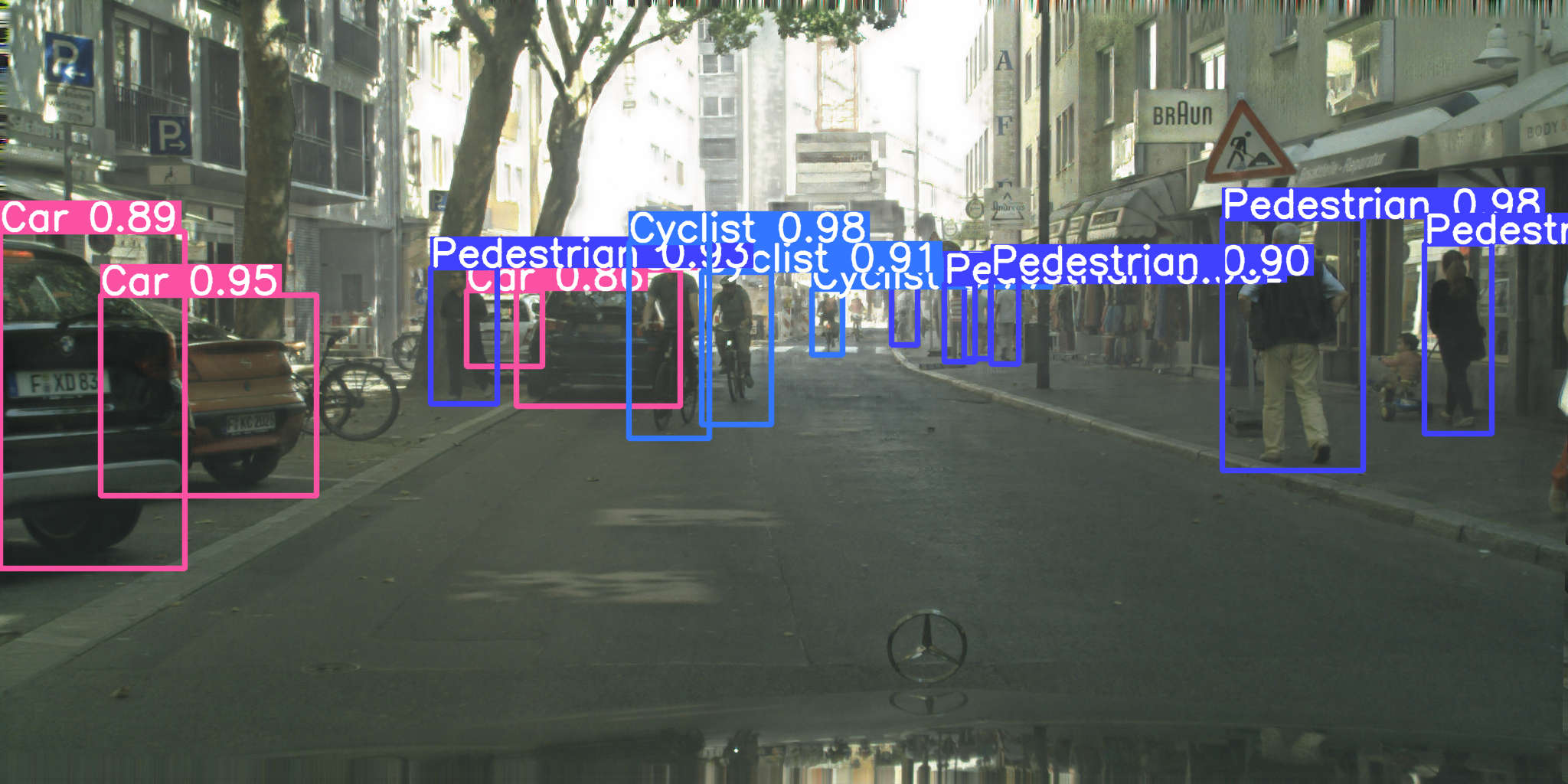}
    \end{minipage}

    \vspace{-0.2cm}
    \begin{minipage}{0.98\linewidth}
        \centering
        \includegraphics[width=\linewidth,height=2.6cm,keepaspectratio]{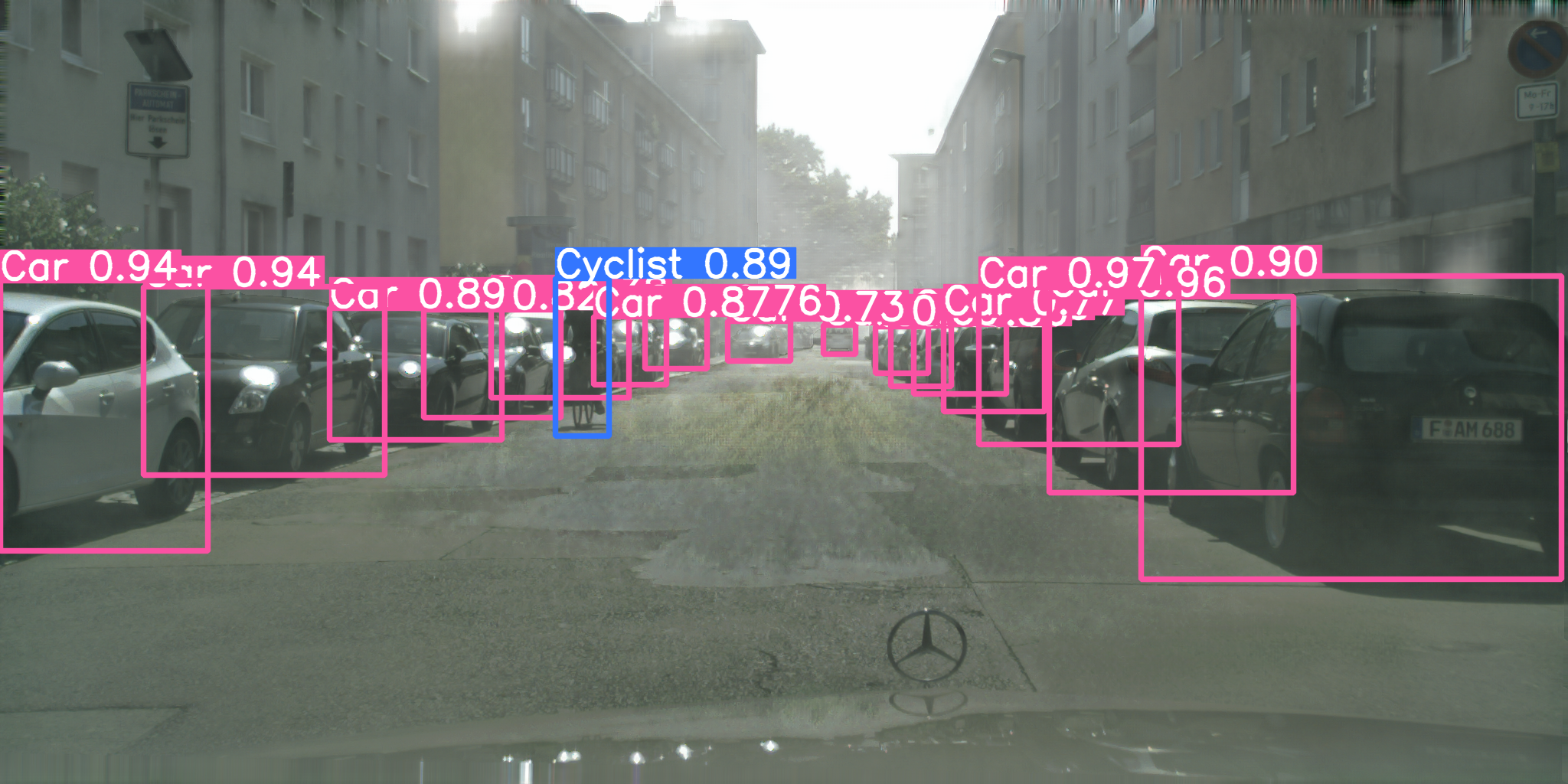}
        \centerline{(c) YOLO-Vehicle-Pro}
    \end{minipage}
\end{minipage}
\end{minipage}
\caption{Performance of the YOLO-Vehicle-Pro model on the Foggy Cityscapes detection test dataset. From left to right, the images show (a) the real-world hazy images, (b) the dehazed images and (c) the detection results on the dehazed image. The YOLO-Vehicle-Pro model adopts a "dehaze-then-detect" strategy, demonstrating excellent detection performance under hazy weather conditions.}
\label{fig:Foggy_citycapes_detection}
\end{figure*}

As observed in Fig.~\ref{fig:yolo_world_haze}, with the decrease in atmospheric visibility, traditional object detection algorithms typically exhibit a significant degradation in performance. However, the YOLO-Vehicle-Pro model demonstrates exceptional robustness and adaptability in hazy conditions. Even under low visibility hazy conditions, the model successfully detects distant vehicles, maintaining high detection accuracy and recall rates. In contrast, the performance of traditional object detection models tends to decline significantly. Fig.~\ref{fig:Foggy_citycapes_detection} shows the performance of the YOLO-Vehicle-Pro model on the Foggy Cityscapes detection test dataset.

\begin{table}[]
\centering
\caption{Performance comparison of the proposed model and previous object detection networks on the Foggy Cityscapes dataset}
\renewcommand{\arraystretch}{1.5}
\begin{tabular}{c|cccc}
\hline
\textbf{Model}            & \textbf{Input    Size} & \textbf{mAP\textsubscript{@50}} & \textbf{FPS} & \textbf{Inference    Time} \\ \hline
YOLOV6-s6        & 2048x1024     & 60.8   & 71  & 50.4ms            \\ \hline
YOLOV8n          & 2048x1024     & 58.4   & 79  & 48.2ms            \\ \hline
YOLOV8s          & 2048x1024     & 61.2   & 73  & 50.1ms            \\ \hline
YOLO-World       & 2048x1024     & 63.2   & 68  & 51.2ms            \\ \hline
YOLO-Vehicle-v1s & 2048x1024     & 65.6   & \textbf{84}  & \textbf{47.7ms}            \\ \hline
\textbf{YOLO-Vehicle-Pro} & 2048x1024     & \textbf{82.3}   & 43  & 76.4ms            \\ \hline
\end{tabular}
\label{tab:performance_comparison}
\end{table}

As shown in the data from Table~\ref{tab:performance_comparison}, the YOLO-Vehicle-Pro model achieves mAP\textsubscript{@50} of 82.3\%, which is 16.7 percentage points higher than the base YOLO-Vehicle model and 21.5 percentage points higher than YOLOV6. These results convincingly demonstrate the superior performance of the YOLO-Vehicle-Pro model in handling hazy scenarios. Notably, YOLO-Vehicle-Pro maintains a high processing speed while preserving high detection accuracy. Although its FPS is slightly lower than the base YOLO-Vehicle model due to the introduction of the dehazing module, its processing speed of 43 FPS still far exceeds traditional two-stage detectors such as Faster R-CNN. To evaluate the performance of the YOLO-Vehicle-Pro model in practical applications, this study also conducted long-term road tests. During a month-long testing period, the model maintained relatively stable detection performance under the hazy weather conditions addressed in this paper.

\subsection{Edge-Cloud Collaborative Intelligent Object Detection System}

In this paper, we propose an innovative edge-cloud collaborative intelligent object detection system aimed at addressing the challenges of real-time performance, accuracy and scalability in autonomous driving scenarios. This system integrates the advantages of edge computing \cite{30},\cite{47} and cloud computing, forming an efficient and flexible intelligent transportation solution. The core of the system is constituted by a mobile platform that integrates various sensors and computing units, ensuring comprehensive environmental perception for vehicles. The specific description content has been moved to the supplementary file.

\section{CONCLUSION}\label{sec_conclusion}
This paper presents the design and implementation of a cloud-edge collaborative object detection system. The system deploys YOLO-Vehicle and YOLO-Vehicle-Pro models on edge devices (such as NVIDIA Jetson Nano) for real-time image acquisition and preliminary processing, while offloading part of the computational tasks to cloud servers under hazy weather conditions or when high-precision detection is required. This cloud-edge collaborative approach fully utilizes the real-time capabilities of edge devices and the powerful computing capacity of the cloud, overcoming the limitations of pure edge computing constrained by hardware resources, while also avoiding network latency issues associated with complete reliance on cloud computing. The core contribution of this research lies in proposing two innovative deep learning models: YOLO-Vehicle and YOLO-Vehicle-Pro. These models are specifically designed to address object detection challenges in environments ranging from clear to low-visibility conditions. YOLO-Vehicle is an object detection model optimized for autonomous driving scenarios, employing multi-scale feature map extraction techniques and region-text feature extraction to achieve efficient and accurate recognition of complex traffic scenes. YOLO-Vehicle-Pro further introduces an improved image dehazing algorithm and adaptive feature extraction mechanism, significantly enhancing detection performance in low-visibility environments. Experimental results demonstrate that on the KITTI dataset, the YOLO-Vehicle-v1s model achieved excellent performance with 92.1\% accuracy, 93.5\% mAP\textsubscript{@50}, and 81.1\% mAP\textsubscript{@75}, while maintaining a detection speed of 226 FPS and an inference time of 12ms. Notably, when processing hazy images, the YOLO-Vehicle-Pro model achieved high precision with 82.3\% mAP\textsubscript{@50} on the Foggy Cityscapes dataset, while maintaining a detection speed of 43 FPS.Furthermore, this paper proposes to apply YOLO-Vehicle and YOLO-Vehicle-Pro to a wider range of edge computing devices and integrate this system into more ITS platforms and scenarios.

\section*{ACKNOWLEDGMENTS}
This work was supported in part by the national natural science foundation of china youth fund 62303331, and in part by the national natural science foundation of china under grant 62372310, and in part by the Liaoning province applied basic research program under grant 2023JH2/101300194, and in part by the Liaoning revitalization talents program under grant XLYC2203151. and in part by the provincial department of education project-basic scientific research operation expenses special fund: Zong 20240176 (Z20240176), and in part by the fundamental research funds for the universities of Liaoning province(LJ222410143095). LLM was used for proofreading.

\bibliographystyle{IEEEtran}
\bibliography{main}

\end{document}